\documentclass[pdflatex,sn-apa]{sn-jnl}

\usepackage{graphicx}%
\usepackage{multirow}%
\usepackage{amsmath,amssymb,amsfonts}%
\usepackage{amsthm}%
\usepackage{mathrsfs}%
\usepackage[title]{appendix}%
\usepackage{xcolor}%
\usepackage{textcomp}%
\usepackage{manyfoot}%
\usepackage{booktabs}%
\usepackage{algorithm}%
\usepackage{algorithmicx}%
\usepackage{algpseudocode}%
\usepackage{listings}%
\usepackage{float}
\usepackage{adjustbox}
\usepackage{multirow}
\usepackage{subcaption}
\usepackage{amsmath,amssymb,amsfonts}
\usepackage{rotating}
\usepackage{arydshln}

\theoremstyle{thmstyleone}%
\theoremstyle{thmstyletwo}%

\theoremstyle{thmstylethree}%

\usepackage{hyperref} 

\newcommand{\orcidurl}[1]{\href{https://orcid.org/#1}{[#1]}}

\begin{document}

\title[An Experimental Study on Fairness-aware Machine Learning]{An Experimental Study on Fairness-aware Machine Learning for Credit Scoring Problems}


\author[1,2]{\fnm{Huyen} \sur{Giang Thi Thu} \orcidurl{0009-0007-6283-3111}} \email{huyengtt@hvnh.edu.vn} 
\author[3]{\fnm{Thang} \sur{Viet Doan}\orcidurl{0009-0009-3072-5532}}\email{thang.dv509@gmail.com}
\author[3]{\fnm{Ha-Bang} \sur{Ban}\orcidurl{0000-0003-2241-5146}}\email{bangbh@soict.hust.edu.vn}
\author*[4]{\fnm{Tai} \sur{Le Quy}\orcidurl{0000-0001-8512-5854}}\email{tailequy@uni-koblenz.de}

\affil[1]{\orgname{Banking Academy of Vietnam}, \orgaddress{\city{Hanoi}, \country{Viet Nam}}}
\affil[2]{\orgname{Vietnam Academy of Science and Technology}, \orgaddress{\city{Hanoi}, \country{Viet Nam}}}
\affil[3]{\orgname{Hanoi University of Science and Technology}, \orgaddress{\city{Hanoi}, \country{Viet Nam}}}
\affil*[4]{\orgname{University of Koblenz}, \orgaddress{\city{Koblenz}, \country{Germany}}}


\abstract{The digitalization of credit scoring has become essential for financial institutions and commercial banks, especially in the era of digital transformation. Machine learning techniques are commonly used to evaluate customers’ creditworthiness. However, the predicted outcomes of machine learning models can be biased toward protected attributes, such as race or gender. Numerous fairness-aware machine learning models and fairness measures have been proposed. Nevertheless, their performance in the context of credit scoring has not been thoroughly investigated. In this paper, we present a comprehensive experimental study of fairness-aware machine learning in credit scoring. The study explores key aspects of credit scoring, including financial datasets, predictive models, and fairness measures. We also provide a detailed evaluation of fairness-aware predictive models and fairness measures on widely used financial datasets. The experimental results show that fairness-aware models achieve a better balance between predictive accuracy and fairness compared to traditional classification models.}




\maketitle
\section{Introduction}
\label{sec:introduction}
The traditional banking system consumes considerable time, requires plenty of human resources, and is tedious to execute. There is a risk that traditional banks will become obsolete due to technological advancements. Therefore, a digital transformation in the banking systems is needed to fasten and ease banking tasks. We have seen the new technologies in the banking sector as vivid examples of digital transformation, such as Robotic Process Automation (RPA), Big Data, Cloud computing, and Blockchain\footnote{\url{https://boostylabs.com/blog/digital-transformation-in-banking}}.
In the digital transformation process, the automation of banking procedures is an essential requirement of banks. Hence, the digitalization of credit scoring is also apparent because credit scoring is a crucial phase in the risk management process of financial organizations and commercial banks. In order to automatically perform customer credit scoring, a variety of machine learning (ML) methods have been applied effectively \citep{bhatore2020machine,dastile2020statistical,dumitrescu2022machine,trivedi2020study}. The experimental results are calculated based on existing customers’ financial and non-financial data at the time of credit scoring and customer rating.

However, apart from the advantages of ML techniques on credit scoring, there is a bunch of evidence regarding the discriminative impact of ML-based decision-making on individuals and groups of people on the basis of protected attributes such as race or gender~\citep{ntoutsi2020bias,le2022survey}. Therefore, ensuring fairness with respect to the protected attributes of ML models is an important requirement.  It is crucial for ML models to be highly accurate while minimizing discrimination against individuals or groups of people with regard to protected attributes.

In the ML research community, fairness-aware ML has been investigated in many domains, such as finance, healthcare, and education~\citep{le2022survey}. However, there are only a few studies on fairness-aware ML in the banking sector, particularly on the credit scoring problem. The pioneering work of \cite{bono2021algorithmic} was the first empirical study of the accuracy and statistical fairness of different credit scoring technologies in the UK context. The experiments were conducted on only a dataset collected in the UK. Then, \cite{kozodoi2022fairness} provided an evaluation of different fairness-aware classifiers on credit scoring datasets. However, they reported experimental results on only three fairness measures. The literature review by \cite{adegoke2024evaluating} offered a thorough analysis of the fairness of credit scoring models in relation to mortgage accessibility for underserved populations. The current work of \cite{mariscal2024implementing} focuses on analyzing the trade-off between performance and fairness of several fairness-aware ML techniques. However, they reported the results on only two fairness measures (demographic parity and equalized odds). Recently, \cite{hurlin2024fairness} presented a framework aimed at formally testing the null hypothesis of fairness and helping lenders and regulatory bodies identify the factors driving unfair outcomes. Nevertheless the experimental results were reported on only the German credit dataset.  

Furthermore, the choice of the fairness criterion has severe consequences for the social impact of lending decisions with credit scoring~\citep{liu2018delayed}. Without constraints, a scoring model leverages all available (including sensitive) information, potentially discriminating against protected groups if doing so improves predictive performance. The goal of incorporating fairness is to modify decision-making (\textit{i.e.}, scoring) practices to achieve equitable, non-discriminatory outcomes. Indeed, more than 20 fairness measures have been introduced in the domain of fairness-aware ML~\citep{verma2018fairness}. Therefore, choosing a suitable fairness measure for the credit scoring problem is not a straightforward circumstance since no metric is universal and fits all circumstances~\citep{mehrabi2021survey,verma2018fairness}. Hence, a comprehensive review and evaluation of fairness-aware ML models and fairness notions on the credit scoring problem is needed. 

In this work, we summarize the prevalent notions of fairness and evaluate the well-known fairness-aware classification models on diverse public credit scoring datasets. Our work makes the following key contributions:
\begin{itemize}
    \item We provide an overview of fairness-aware ML and prevalent fairness measures applicable to the credit scoring problem.
    \item We analyze popular credit scoring datasets using Bayesian networks and data analytics.
    \item We present a comprehensive evaluation of traditional and fairness-aware classification models on credit scoring datasets.
\end{itemize}

This paper is structured as follows: Section~\ref{sec:fairness_ML} provides an introduction to fairness-aware ML techniques and fairness-aware ML models that can be used for the credit scoring problem. Section~\ref{sec:fairness_measures} describes the most popular fairness measures used in fairness-aware ML models. The following section~\ref{sec:dataset} demonstrates an overview of the datasets used for credit scoring. Next, section~\ref{sec:evaluation} evaluates fairness-aware ML models with fairness notions and credit scoring results from the predictive models. Finally, we outline the conclusions and present some possible future research directions in Section~\ref{sec:conclusion}.

\section{Fairness-aware Machine Learning (for Credit Scoring)}
\label{sec:fairness_ML}
In this section, we provide an overview of fairness-aware predictive models designed for classification tasks in the financial domain, with a particular focus on models potentially applicable to the credit scoring problem. We review three main categories of fairness-aware predictive models: pre-processing, in-processing, and post-processing.

\subsection{Formulation of the Credit Scoring Problem}
\label{subsec:problem}
The credit scoring problem is described as ``methods used for classifying applicants for credit into ‘good’ and ‘bad’ risk classes'' \citep{hand1997statistical,dastile2020statistical}. Similarly, it can be defined as a ``set of decision models and their underlying techniques that aid credit lenders in the granting of credit'' \citep{thomas2017credit} or a tool used to quantify credit risk using applicants’ financial behavior and repayment history~\citep{ayari2025machine}. Therefore, in this paper, we consider the credit scoring problem as a binary classification problem.

We denote $D$ as a dataset with class attribute $Y = \{+, -\}$; e.g., $Y = \{good\phantom{a}credit, bad\phantom{a}credit\}$ or $Y = \{accepted, rejected\}$, etc. A binary protected attribute is denoted by $S$, $S \in \{s,\overline{s}\}$ where $s$ is the protected group and $\overline{s}$ is the non-protected group; e.g., $S = $ \emph{``Sex''} and $S \in \{female, male\}$.  $\hat{Y} =\{+, –\}$ is the predicted class. Hence, the protected and non-protected groups with respect to positive (negative, respectively) classes are $s_{+}$ ($s_{-}$), $\overline{s}_{+}$ ($\overline{s}_{-}$). We refer to the positive class as the target class, e.g., \textit{good credit}. 

The goal of the fairness-aware classification model in the credit scoring problem is to find a map function $f: D \mapsto Y$ that minimizes the loss and mitigates the discriminatory outcomes simultaneously.

\subsection{Fairness-aware ML Models}
\label{subsec:fairmethods}
There are three approaches to mitigating bias in ML models and achieving fairness: i) pre-processing methods; ii) in-processing methods; and iii) post-processing methods~\citep{mehrabi2021survey,ntoutsi2020bias}.

In the pre-processing approach, researchers focus on the data, which are the primary source of bias. The goal is to generate a ``balanced'' dataset and then apply any ML algorithms to that. For example, the class labels are altered, different weights are assigned to instances, or the protected and unprotected groups are balanced in the training set.  Techniques such as learning fair representations (LFR) aim to encode data effectively while obscuring protected attributes~\citep{zemel2013learning}. Similarly, the disparate impact remover (DIR) adjusts feature values to enhance group fairness while preserving rank-ordering within groups~\citep{feldman2015certifying}.

In-processing approaches reformulate the classification problem by explicitly incorporating the model’s discrimination behavior in the objective function through regularization or constraints or by training on latent target labels. Besides, an in-processing approach involves incorporating a model’s discrimination behavior into the objective function by regularizing or constraining it. According to Agarwal’s method, a fair classification can be reduced to a series of cost-sensitive classification problems with the lowest (empirical) error under the desired constraints~\citep{agarwal2018reductions}. AdaFair~\citep{iosifidis2019adafair}, a sequential fair ensemble, extends AdaBoost’s weighted distribution approach by taking into account the cumulative fairness of the learner up until the current boosting round and moreover, accounts for class imbalance by optimizing for balanced error instead of an overall error.

Unlike the above two approaches, the post-processing method post-process the classification models once they have been learned from data. It involves altering the model’s internals (white-box approaches) or its predictions (black-box approaches). White-box post-processing methods adjust the internal decision-making criteria of a model~\citep{kamiran2012data}. For example, decision thresholds might be altered to balance outcomes across sensitive groups. This requires direct access to the model's decision rules, making it most suitable for scenarios where model transparency is available. Black-box post-processing methods, by contrast, operate solely on the model's outputs~\citep{kim2019multiaccuracy}. This makes them model-agnostic and widely applicable. For instance, calibrated equalized odds post-processing (CEP) optimizes calibrated classifier score outputs to determine the probabilities of altering output labels to achieve equalized odds~\citep{pleiss2017fairness}. Similarly, equalized odds post-processing (EOP) uses linear programming to find probabilities for modifying output labels, ensuring equalized odds objectives are met~\citep{hardt2016equality}.

In this work, we demonstrate the performance of the three above approaches with 6 well-known models: i) Pre-processing approach: Learning fair representations (LFR), Disparate impact remover (DIR); ii) In-processing approach: Agarwal’s, AdaFair; iii) Post-processing approach: Equalized odds post-processing (EOP), Calibrated equalized odds post-processing (CEP).

\section{Fairness Measures}
\label{sec:fairness_measures}

We perform the evaluation on the most popular group fairness notions which are used to determine how fair the model's results are. The fairness notion may be turned into measures by taking a difference or a ratio of the equation components~\citep{vzliobaite2017measuring}. Therefore, in this paper, we use the terms ``fairness notion'' and ``fairness measure'' interchangeably. The fairness measures are chosen based on the number of citations\footnote{Reported by Google Scholar on 21st October 2024}. In all fairness measures, a higher value indicates a larger difference in predictions between the two groups, so the model is less fair, i.e., 0 stands for no discrimination. Table~\ref{tbl:measures} provides an overview of fairness measures. Fairness measures are defined as below using notations described in Section~\ref{subsec:problem}.

\begin{table}[!h]
\resizebox{\textwidth}{!}
{
\begin{minipage}{\textwidth}
\centering
\caption{An overview of fairness measures}
\label{tbl:measures}

    \begin{tabular}{ l  c c} \hline 
        \textbf{Fairness measures}  & \textbf{\#Citations} &  \textbf{Values} \\ \hline
         Statistical parity (SP)  \citep{dwork2012fairness} & 4398  & $[-1, 1]$\\  
         Equal opportunity (EO)   \citep{hardt2016equality} & 4935 & $[0, 1]$\\  
         Equalized odds (EOd)     \citep{hardt2016equality} & 4935 & $[0, 2]$\\  
         Predictive parity (PP)   \citep{chouldechova2017fair} & 2461 & $[0, 1]$\\  
         Predictive equality (PE) \citep{corbett2017algorithmic} & 1545 & $[0, 1]$\\  
         Treatment equality (TE)  \citep{berk2021fairness} & 1170 & $(-\infty, \infty)$\\ 
         ABROCA                   \citep{gardner2019evaluating}& 191  & $[0, 1]$\\ \hline
    \end{tabular}       
\end{minipage}}
\end{table}

\textbf{Statistical parity (SP)}
\begin{equation}
\label{eq:statistical_parity}
    SP =  P(\hat{Y}=+ \mid S=\overline{s}) - P(\hat{Y}=+ \mid S=s)    
\end{equation}

\textbf{Equal opportunity (EO)}
\label{eq:equal_opportunity}
\begin{equation}
    EO =\mid P(\hat{Y} = - \mid Y = +, S = \overline{s}) - P(\hat{Y} = - \mid Y = +, S=s)\mid
\end{equation}  

\textbf{Equalized odds (EOd)}
\begin{equation}
\label{eq:equalized_odds}
EOd = \sum_{y\in \{+,-\}}\mid P(\hat{Y}=+ \mid S=s,Y=y) - P(\hat{Y}=+\mid S=\overline{s},Y=y)\mid
\end{equation}

\textbf{Predictive parity (PP)}
\begin{equation}
\label{eq:predictive_parity}
    PP = \mid P(Y = + \mid \hat{Y} = +, S = s) - P(Y = + \mid\hat{Y} = +, S = \overline{s})\mid
\end{equation}

\textbf{Predictive equality (PE)}
\label{eq:predictive_equality}
\begin{equation}
PE = \mid P(\hat{Y} = + \mid Y = -,S = s) - P(\hat{Y} = + \mid Y = -,S = \overline{s})\mid
\end{equation}

\textbf{Treatment equality (TE)} Treatment equality is computed based on False Negative (FN) and False Positive (FP) of the protected group (prot.) and non-protected (non-prot.) groups.
\begin{equation}
\label{eq:treatment_equality}
    \frac{FN_{prot.}}{FP_{prot.}} = \frac{FN_{non-prot.}}{FP_{non-prot.}}
\end{equation}

\textbf{Absolute Between-ROC\footnote{ROC: Receiver operating characteristic} Area (ABROCA)}
It measures the divergence between the protected ($ROC_s$) and non-protected group ($ROC_{\overline{s}}$) curves across all possible thresholds $t \in [0, 1]$ of false positive rates (FPR) and true positive rates (TPR). The absolute difference between the two curves is calculated to account for cases where the curves intersect.

\begin{equation}
\label{eq:abroca}
    \int_{0}^{1}\mid ROC_{s}(t) - ROC_{\overline{s}}(t)\mid \,dt
\end{equation}

\section{Datasets for Credit Scoring}
\label{sec:dataset}
This section provides a systematic view of financial datasets used for the credit scoring problem. We perform fundamental analysis to discover bias in the dataset itself by analyzing the association of protected attributes with class attributes.

\begin{figure}
    \centering
    \includegraphics[width=0.9\linewidth]{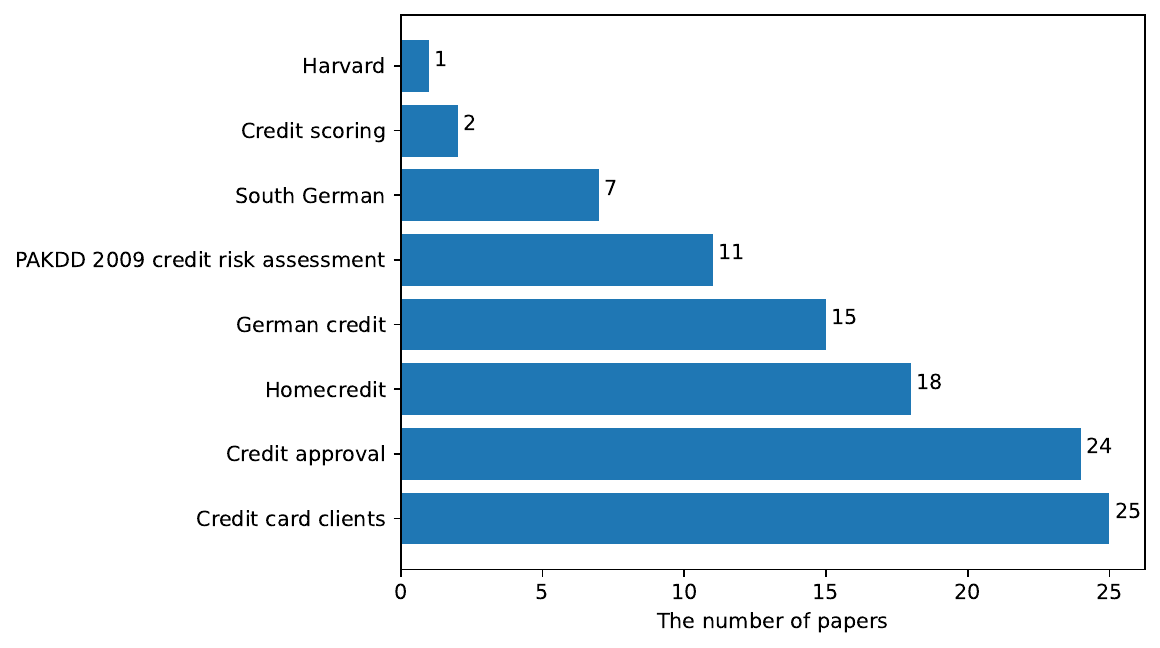}
    \caption{The use of credit scoring datasets}
    \label{fig:datasetcount}
\end{figure}

To identify the relevant datasets, we use several research databases such as Google Scholar\footnote{\url{https://scholar.google.com/}}, Paper With Code\footnote{\url{https://paperswithcode.com/}}, ResearchGate\footnote{\url{https://www.researchgate.net/}}, ScienceDirect\footnote{\url{https://www.sciencedirect.com/}}
with ``datasets for credit scoring'' as the primary query term to narrow down the search. We take into account the resulting papers from 2010 to 2021 because this was the post-global recession period \citep{mcdonald2009global,kose2020global}, and credit lending became a challenging issue due to the emergence of various inequalities and a lack of transparency in credit activities. Figure~\ref{fig:datasetcount} illustrates the use of found datasets in scientific works.
We select datasets for our experiments based on several criteria: i) The dataset must contain the protected attributes, such as gender, race, etc.; ii) The dataset must have the ``target/class'' attribute which is used for classification operation; iii) The dataset must have more than 500 instances. To this end, we employ 5 datasets for our evaluation\footnote{We use the term ``Sex'' to refer to ``Gender'', ``Marital status'' to refer to ``Marriage'', ``Family status''. Abbreviation: IR: Imbalance ratio.}, described in Table~\ref{tbl:datasets}.

\begin{table}[!h]  
\resizebox{\textwidth}{!}
{
\begin{minipage}{\textwidth}
\centering
\caption{An overview of credit scoring datasets}
\label{tbl:datasets}
\begin{tabular}{lrrcccc}
\hline
\multicolumn{1}{c}{\textbf{ Datasets }} &  
\multicolumn{1}{c}{\textbf{ \#Instances}} &
\multicolumn{1}{c}{\begin{tabular}[c]{@{}c@{}}\textbf{ \#Instances}\\\textbf{  (cleaned)} \end{tabular}} &
\multicolumn{1}{c}{\textbf{ \#Attributes }} &
\multicolumn{1}{c}{\begin{tabular}[c]{@{}c@{}}\textbf{ Protected }\\\textbf{ attribute(s) } \end{tabular}} &
\multicolumn{1}{c}{\begin{tabular}[c]{@{}c@{}}\textbf{ Class label }\\\textbf{ (positive) } \end{tabular}} & 
\textbf{ IR (+:-) } \\ \hline
Credit approval & 690    & 678       & 15 & Sex, Age                 & Approved           & 1:1.23  \\
Credit card clients& 30,000 & 30,000    & 23 & Sex, Education, Marital status & Default payment    & 1:3.52  \\
Credit scoring  & 8,755  & 8,755     & 17 & Age, Sex, Marital status        & Good credit        & 11.58:1 \\
German credit   & 1,000  & 1,000     & 21 & Age, Sex                 & Good credit        & 2.33:1  \\
PAKDD credit    & 50,000 & 38,896    & 47 & Age, Sex, Marital status & Bad credit         & 2.83:1  \\
\hline
\end{tabular}
\end{minipage}}
\end{table}

In the next step, inspired by the work of \cite{le2022survey}, we perform fundamental data analysis to investigate bias in the dataset by using the Bayesian network (BN) \citep{holmes2008innovations}. The BN is used to discover the relationship between protected attributes and class label. If the generated BN reveals any direct or indirect connection between a protected attribute and the class attribute, we can infer that the dataset may be biased with respect to that specific protected attribute. We also transform the numerical attributes into categorical attributes in order to reduce the computation complexity of the BN generator \citep{chen2017learning}. Regarding the BN of the \emph{Credit card clients} and \emph{German credit} datasets, we refer to the work of \cite{le2022survey}.

\subsection{Credit Approval Dataset}
\begin{figure}
    \centering
    \includegraphics[width=1\linewidth]{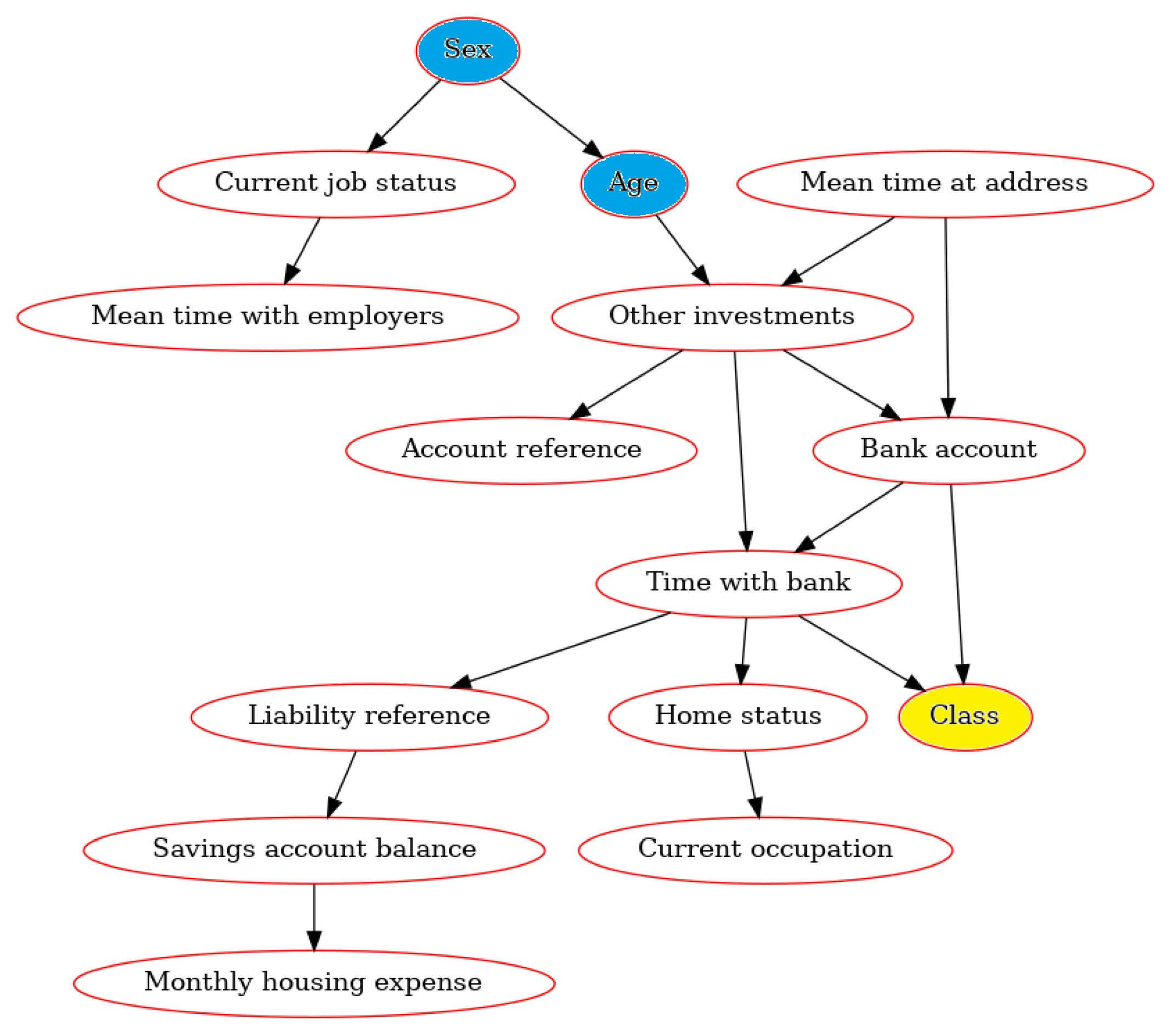}
    \caption{Credit approval: Bayesian network (class label: \emph{Class}, protected attributes: \emph{Age, Sex}).}
    \label{fig:BN-credit-approval}
\end{figure}

\begin{figure}
    \centering
    \includegraphics[width=0.5\linewidth]{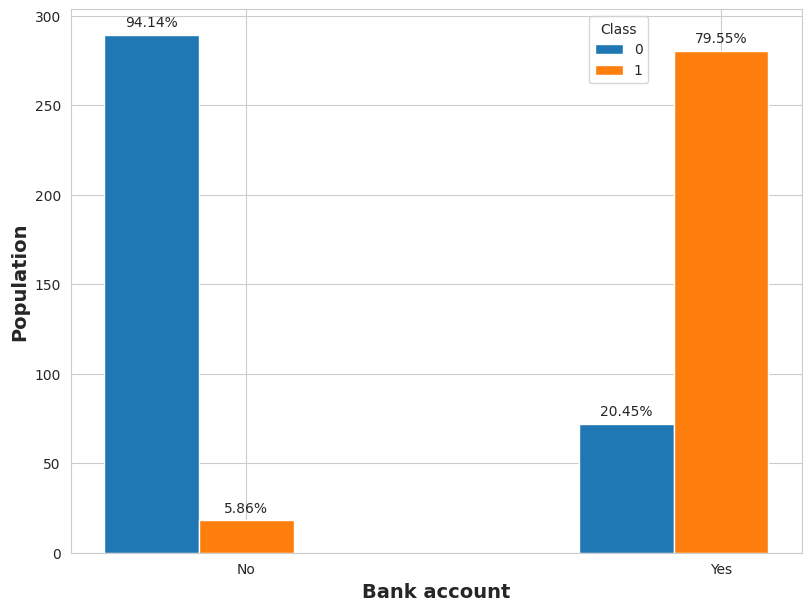}
    \caption{Credit approval: Relationship between class label and \emph{Bank account} attribute.}
    \label{fig:credit-approval-class-bank}
\end{figure}

The credit approval dataset\footnote{\url{http://archive.ics.uci.edu/ml/datasets/credit+approval}}  (another name: Australian credit approval dataset) contains information of 690 credit card applications. The classification task is to predict whether an application is approved or not (class attribute: \textit{Approved}). The positive class is approved (value +). To generate the BN, we discretize $age = \{<25, \ge25\}$; the continuous variables \textit{Debt}, \textit{YearsEmployed}, \textit{CreditScore} and \textit{Income} are encoded based on their median value: $Debt = \{\le 2.875, >2.875\}$, \textit{YearsEmployed} =$\{\le 1, >1\}$, \textit{Income} =$\{\le 5, >5\}$. Figure \ref{fig:BN-credit-approval} depicts the BN of the credit approval dataset where the class label is highlighted in yellow, while the protected attributes are colored in blue. In the BN, there is a strong relationship between \emph{Class} and \emph{Bank account} attributes. The analysis shows that 79.55\% of people with bank accounts (\emph{Bank account = ``Yes'')} are approved for credit, while the rate among people without bank accounts is only 5.86\% (Figure \ref{fig:credit-approval-class-bank}).

\subsection{Credit Card Clients Dataset} 
The credit card clients dataset\footnote{\url{https://archive.ics.uci.edu/dataset/350/default+of+credit+card+clients}} consists of information about 30,000 customers in Taiwan in October 2005 \citep{yeh2009comparisons}. The prediction task is to forecast if a customer will face the default situation in the next month or not (class attribute: \textit{Y}). The positive class is default payment (value 0). 

\subsection{Credit Scoring Dataset}
\begin{figure}
    \centering
    \includegraphics[width=1\linewidth]{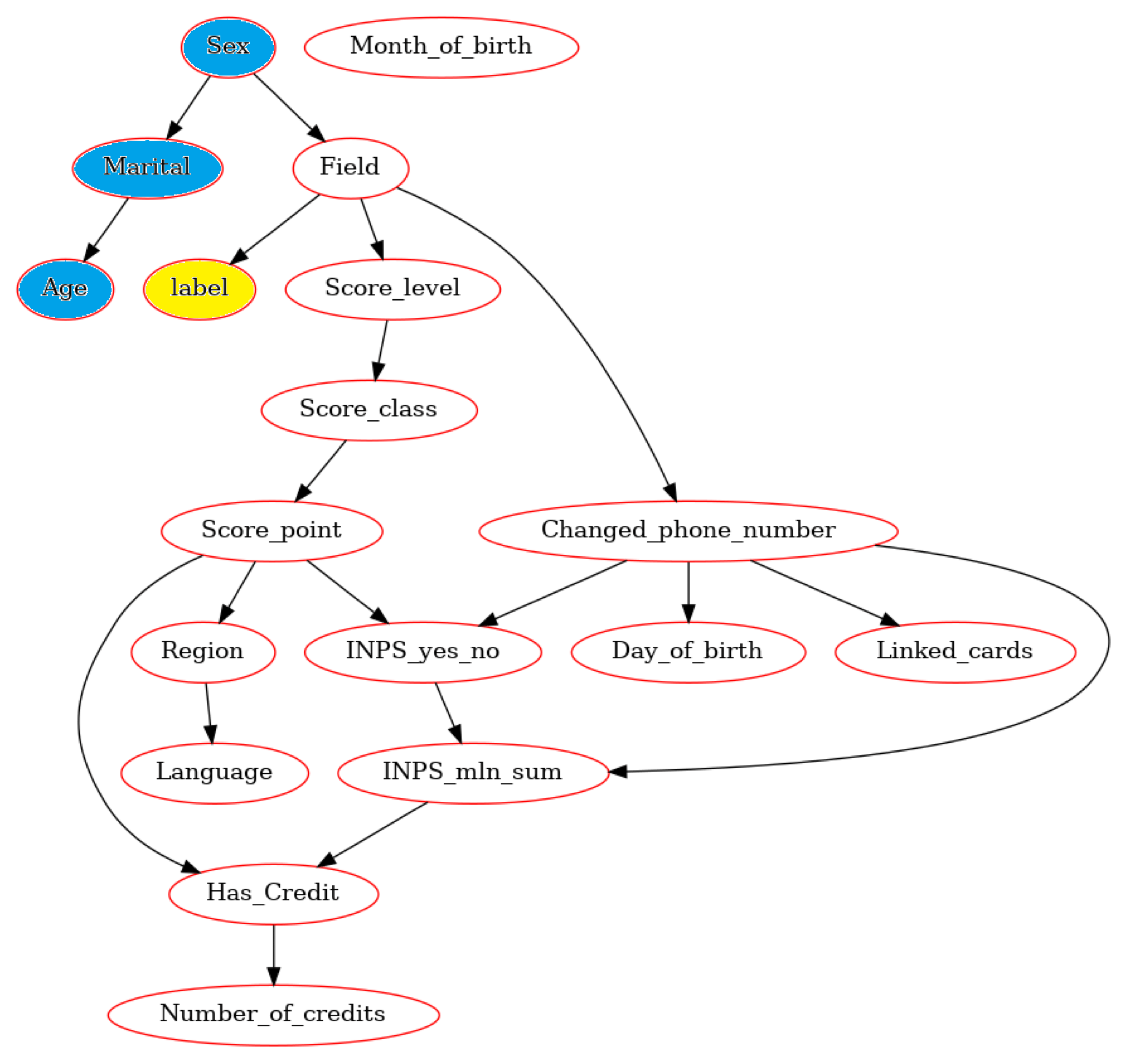}
    \caption{Credit scoring: Bayesian network (class label: \emph{Label}, protected attributes: \emph{Age, Marital, Sex}.}
    \label{fig:BN-credit-scoring}
\end{figure}

The credit scoring dataset\footnote{\url{https://www.kaggle.com/code/islombekdavronov/credit-scoring}} has 8,755 records of customers collected by a FinTech company in Central Asia. The dataset was published on Kaggle in 2021 by Davronov. Predicting whether a customer has good credit (value 1) is the main goal (class attribute: \textit{label}). We categorize two numerical attributes based on their median value: \emph{INPS\_mln\_sum =\{ $\le 1.7, > 1.7$\}} and \emph{Score\_point = \{$\le 0, > 0$\}}. Figure \ref{fig:BN-credit-scoring} demonstrates the BN of the Credit scoring dataset. There is an indirect relationship between the class \emph{label} and \emph{Sex} attribute which might imply a bias in the dataset.

\subsection{German Credit Dataset}
The German credit dataset\footnote{\url{https://archive.ics.uci.edu/dataset/144/statlog+german+credit+data}} contains observations for 1000 applicants for credit. It was published on the UCI repository website by \citet{statlog_(german_credit_data)_144}. The goal is to predict whether a customer has good (value 1) or bad credit (value 2) (class attribute: \textit{class-label}). The positive class is good (value 1).
\subsection{PAKDD 2009 Credit Risk Assessment Dataset}
\begin{figure}
    \centering
    \includegraphics[width=1\linewidth]{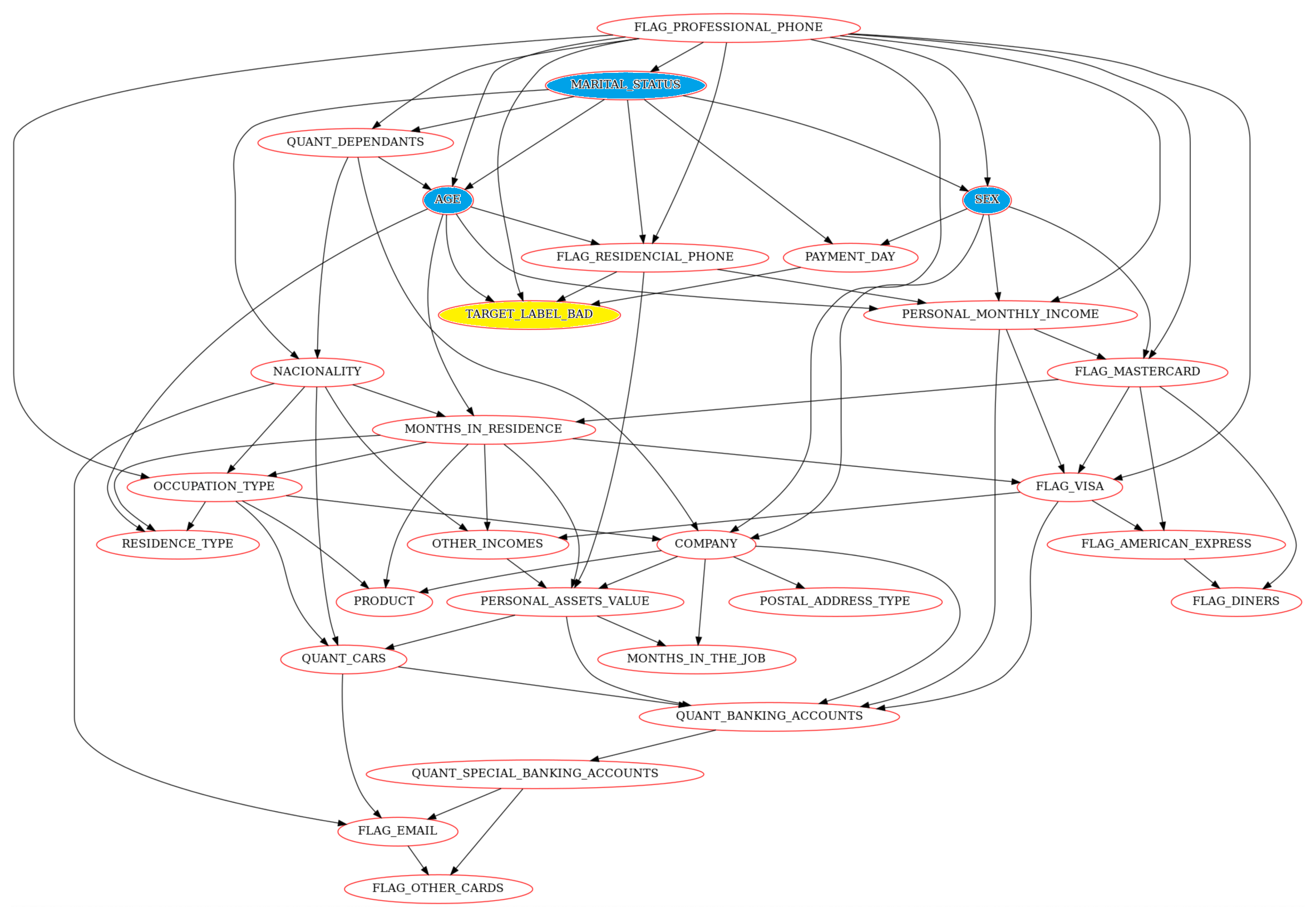}
    \caption{PAKDD credit: Bayesian network (class label: \emph{TARGET\_LABEL\_BAD}, protected attributes: \emph{AGE, SEX, MARITAL\_STATUS}).}
    \label{fig:BN-PAKDD}
\end{figure}

\begin{figure}
    \centering
    \includegraphics[width=0.8\linewidth]{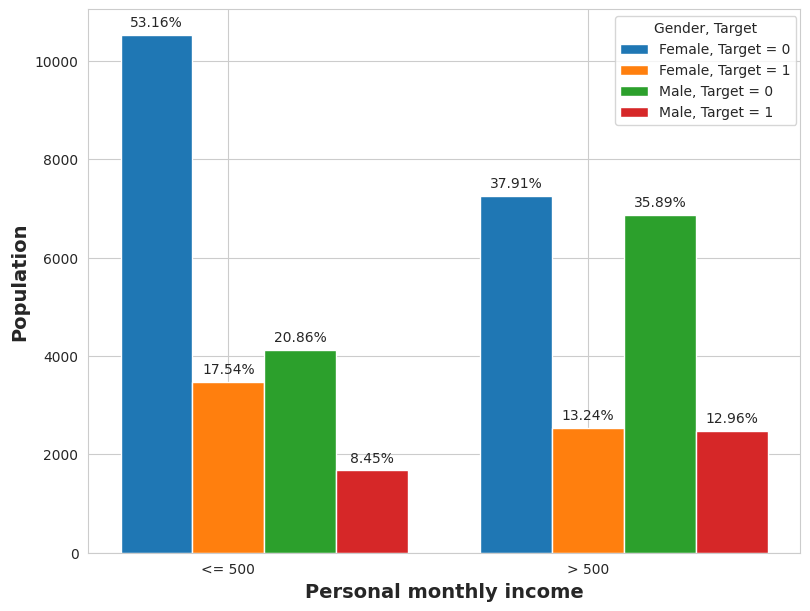}
    \caption{PAKDD credit: Relationship between \emph{Sex, Personal\_monthly\_income} and class label.}
    \label{fig:PAKDD-analysis}
\end{figure}

The PAKDD credit risk assessment dataset\footnote{\url{https://github.com/JLZml/Credit-Scoring-Data-Sets}} was provided by the PAKDD data mining competition in 2009 with 50,000 instances. The class label is \textit{TARGET\_LABEL\_BAD}, intending to predict if a customer has bad credit. Therefore, the positive class in this dataset is set based on the value \textit{TARGET\_LABEL\_BAD = 1} (bad credit). To generate the BN, we remove attributes with too many distinct values, such as \emph{Id\_client, Professional\_code, Residencial\_phone\_area\_code}. Moreover, we discrete continuous attributes based on their median: \emph{Quant\_dependants =\{$\le 0, > 0$\}; Months\_in\_residence =\{$\le 5, > 5$\}; Personal\_monthly\_income = \{$\le 500, > 500$\}}. Figure \ref{fig:BN-PAKDD} depicts the BN of the PAKDD credit dataset. \emph{Sex} attribute has an indirect connection with the class label. We observe that up to 53.16\% of female customers have bad credit ratings when their monthly income is low (below 500). Nevertheless, even among female customers with good credit ratings, the proportion in the higher income group (above 500) remains not high (Figure \ref{fig:PAKDD-analysis}).

\section{Evaluation}
\label{sec:evaluation}
We experiment with selected fairness-aware predictive models using prevalent fairness measures and various financial datasets to evaluate the performance of a wide range of predictive models (traditional models, pre-processing, in-processing, and post-processing fairness-aware predictive models). The preliminary results are the primary means of selecting the appropriate fairness-aware predictive models and fairness measures for credit scoring problem.
\subsection{Experimental Setups}
\label{subsec:setup}
\subsubsection{Predictive Models}
\label{subsubsec:models}
\textbf{Traditional predictive models.} We perform experiments on four well-known traditional classification models applied for the credit scoring problem \citep{brown2012experimental,trivedi2020study}, namely Decision Tree (DT); Naive Bayes (NB); Multi-layer Perceptron (MLP); and k-nearest neighbors (kNN).

\textbf{Fairness-aware ML models.} Three groups of fairness-aware ML models are chosen: i) Pre-processing approach: Learning fair representations (LFR)~\citep{zemel2013learning}, Disparate impact remover (DIR)~\citep{feldman2015certifying}; ii) In-processing approach: Agarwal's \citep{agarwal2018reductions}, Adafair \citep{iosifidis2019adafair}; iii) Post-processing approach: Calibrated equalized odds post-processing (CEP)~\citep{pleiss2017fairness}, Equalized odds post-processing (EOP)~\citep{hardt2016equality}.

\subsubsection{Training and Test Sets}
\label{subsubsec:train-test}
We use 70\% of the data for training and 30\% for testing with a single split. All traditional predicted models are implemented and executed with default parameters provided by the Scikit-learn toolkit\footnote{\url{https://scikit-learn.org/}}. Regarding fairness-aware ML models, we use the implementation of \cite{iosifidis2019adafair} and the AI Fairness 360 toolkit\footnote{\url{https://github.com/Trusted-AI/AIF360}} to execute Agarwal’s, LFR, DIR, EOP and CEP methods. In addition, we combine the use of pre-processing and post-processing models with the traditional approaches. In detail, the resulting datasets of pre-processing models, i.e., fair dataset, will be the input of traditional models. Similarly, the outcome of traditional models will be processed by the post-processing fairness-aware models in order to mitigate bias and achieve fairness in the final output. \emph{Sex} is selected as the protected attribute for all datasets due to its popularity.

We report the prediction performance of classification models for each dataset in the F1 score and balanced accuracy (BA) measures because most datasets are imbalanced, as demonstrated in the imbalance ratio (IR) column of Table \ref{tbl:datasets}.

\subsection{Experimental Results}
\label{subsec:results}

\subsubsection{Credit Approval Dataset}
\label{subsubsec:result-credit-approval}
\begin{figure*}[!h]
\centering
\begin{subfigure}{.30\linewidth}
    \centering
    \includegraphics[width=\linewidth]{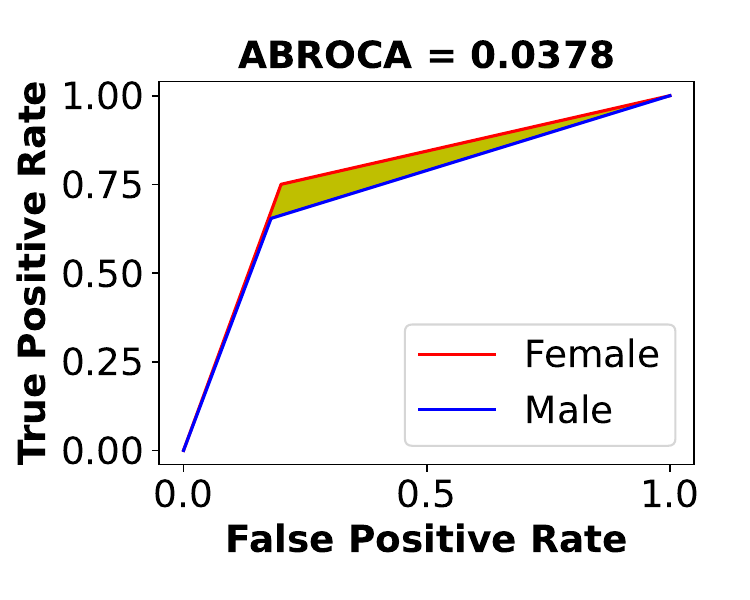}
    \caption{DT}
\end{subfigure}
\hfill
\vspace{-5pt}
\begin{subfigure}{.30\linewidth}
    \centering
    \includegraphics[width=\linewidth]{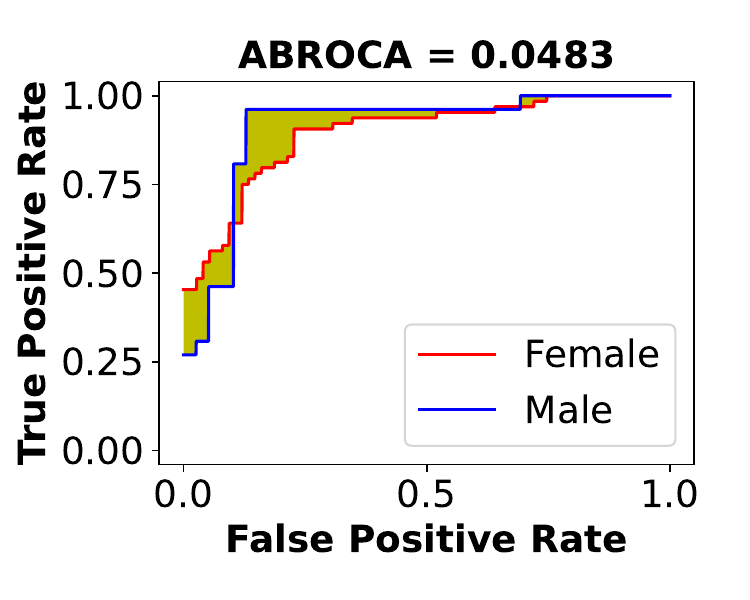}
    \caption{NB}
\end{subfigure}    
\hfill
\vspace{-5pt}
\begin{subfigure}{.30\linewidth}
    \centering
    \includegraphics[width=\linewidth]{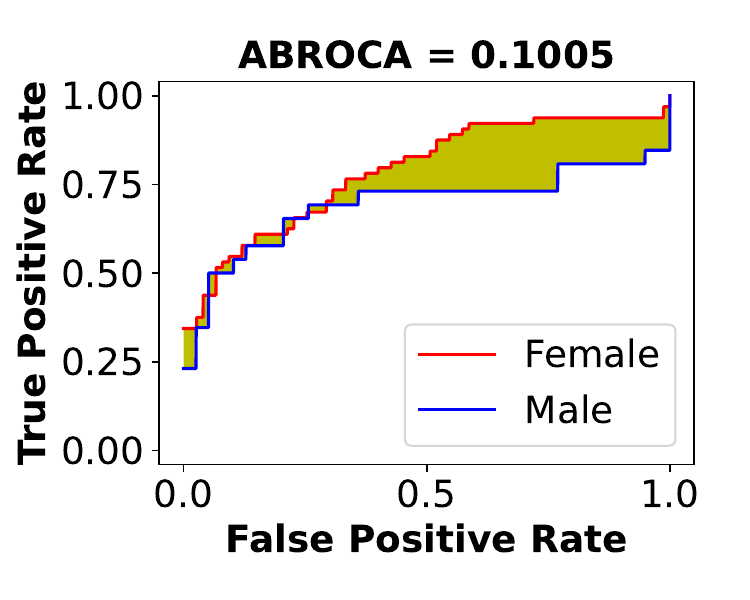}
    \caption{MLP}
\end{subfigure}
\bigskip
\vspace{-5pt}
\begin{subfigure}{.30\linewidth}
    \centering
    \includegraphics[width=\linewidth]{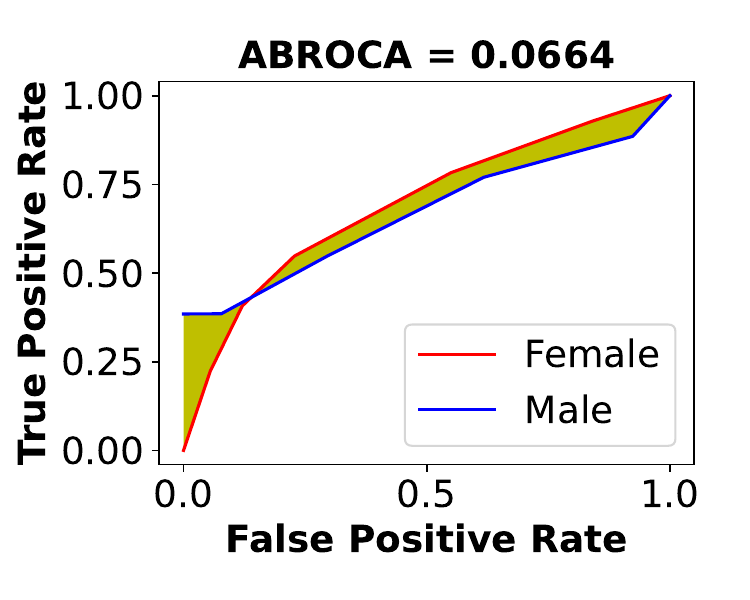}
    \caption{kNN}
\end{subfigure}
\hfill
\vspace{-5pt}
\begin{subfigure}{.30\linewidth}
    \centering
    \includegraphics[width=\linewidth]{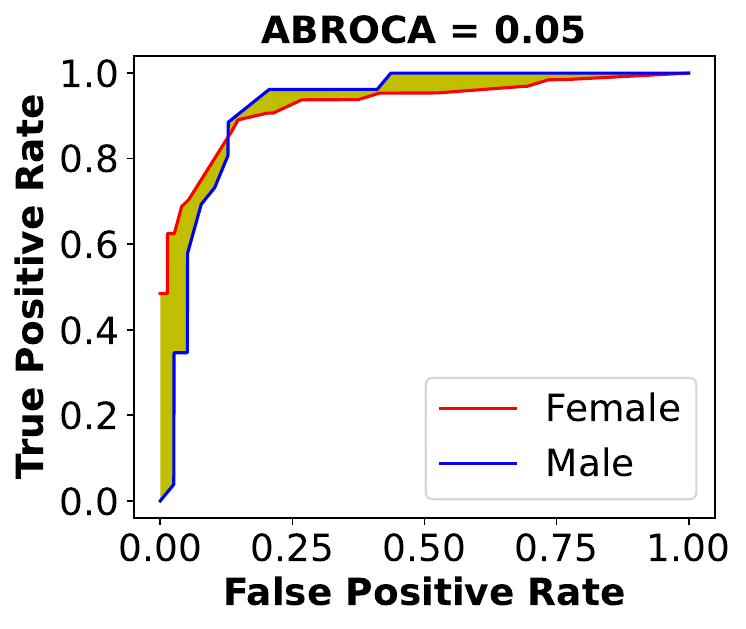}
    \caption{AdaFair}
\end{subfigure}
\vspace{-5pt}
 \hfill
\begin{subfigure}{.30\linewidth}
    \centering
    \includegraphics[width=\linewidth]{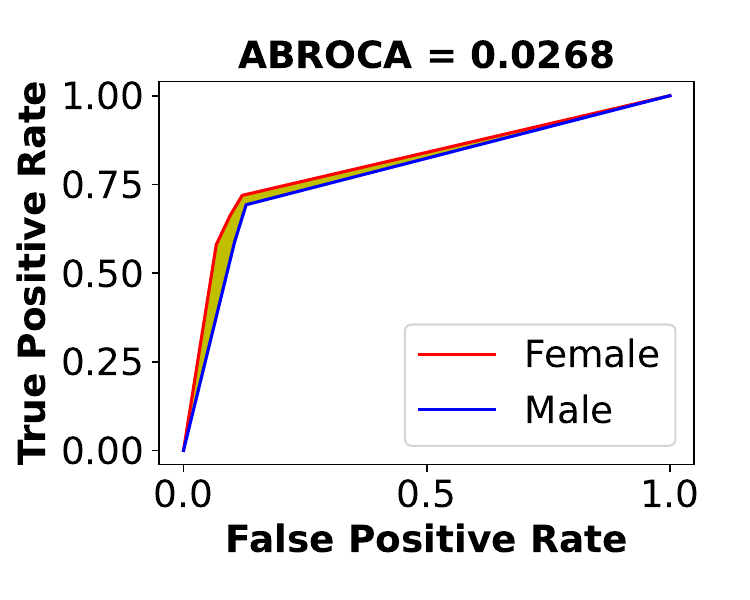}
    \caption{Agarwal's}
\end{subfigure}
\bigskip
\vspace{-5pt}
\begin{subfigure}{.30\linewidth}
    \centering
    \includegraphics[width=\linewidth]{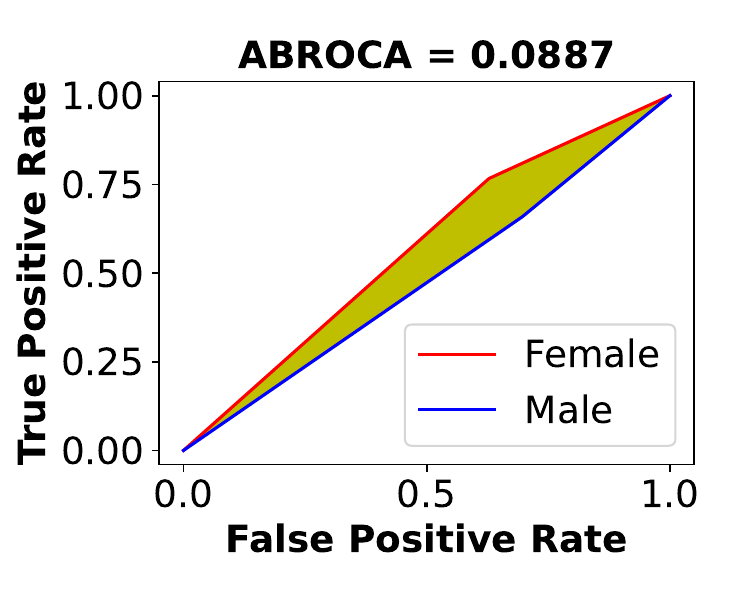}
    \caption{DIR-DT}
\end{subfigure}
\hfill
\vspace{-5pt}
\begin{subfigure}{.30\linewidth}
    \centering
    \includegraphics[width=\linewidth]{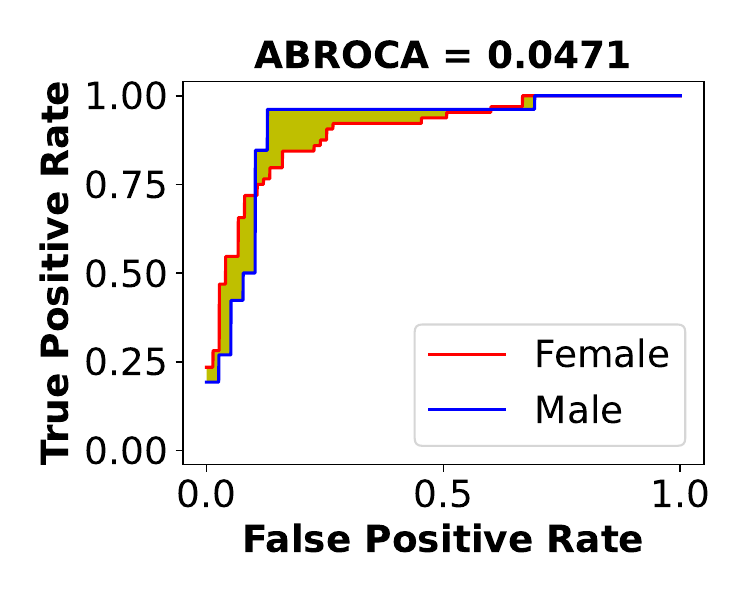}
    \caption{DIR-NB}
\end{subfigure}
\vspace{-5pt}
 \hfill
\begin{subfigure}{.30\linewidth}
    \centering
    \includegraphics[width=\linewidth]{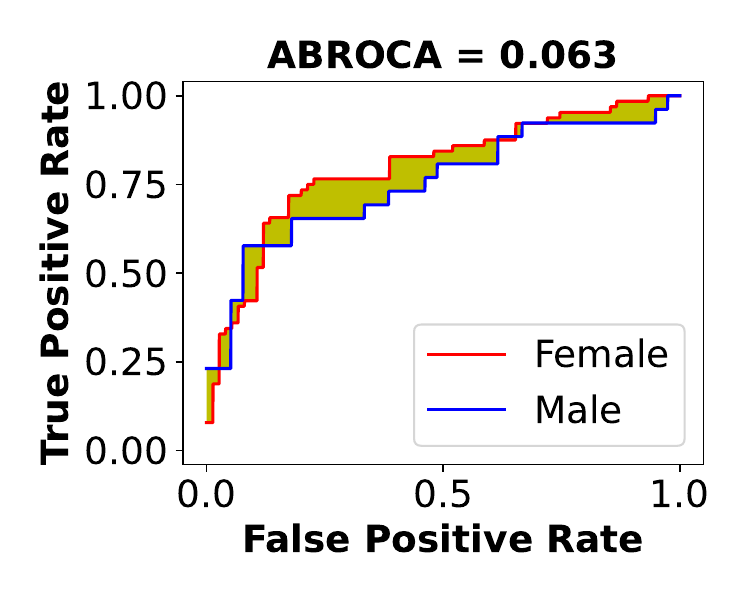}
    \caption{DIR-MLP}
\end{subfigure}
\bigskip
\vspace{-5pt}
\begin{subfigure}{.30\linewidth}
    \centering
    \includegraphics[width=\linewidth]{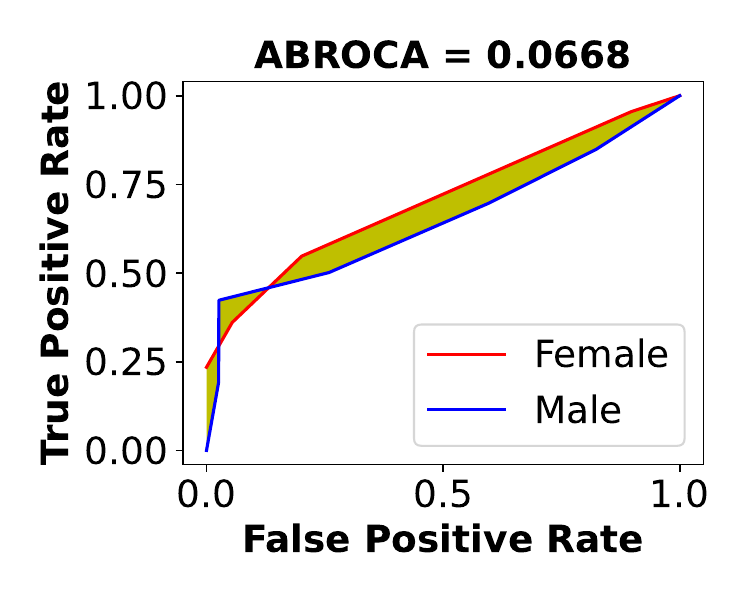}
    \caption{DIR-kNN}
\end{subfigure}
\hfill
\vspace{-5pt}
\begin{subfigure}{.30\linewidth}
    \centering
    \includegraphics[width=\linewidth]{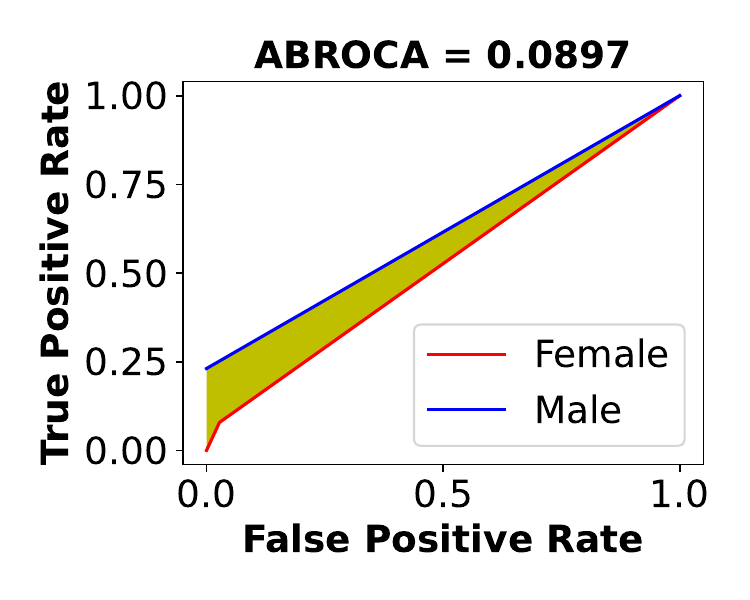}
    \caption{LFR-DT}
\end{subfigure}
\vspace{-5pt}
 \hfill
\begin{subfigure}{.30\linewidth}
    \centering
    \includegraphics[width=\linewidth]{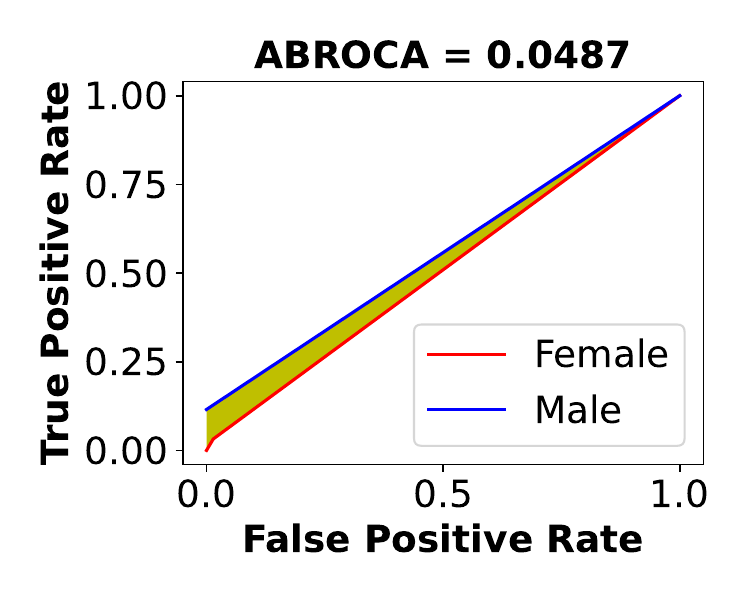}
    \caption{LFR-NB}
\end{subfigure}
\bigskip
\vspace{-5pt}
\begin{subfigure}{.30\linewidth}
    \centering
    \includegraphics[width=\linewidth]{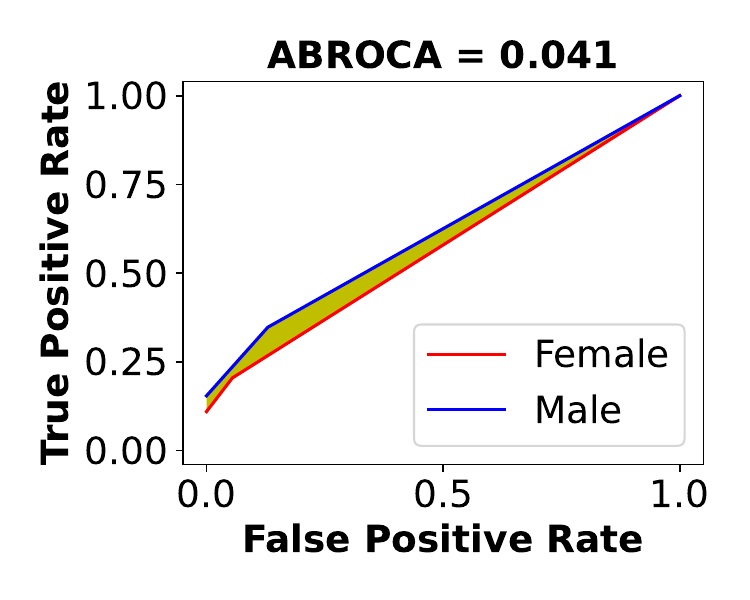}
    \caption{LFR-MLP}
\end{subfigure}
\vspace{-5pt}
\begin{subfigure}{.30\linewidth}
    \centering
    \includegraphics[width=\linewidth]{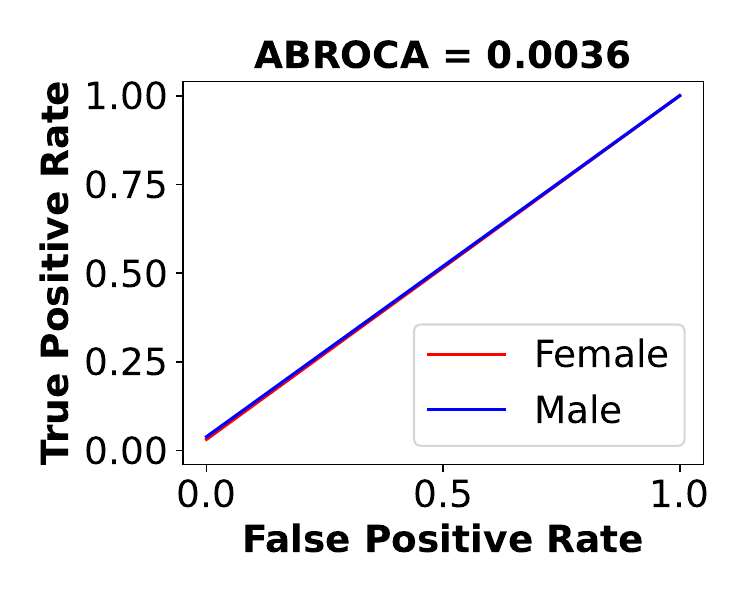}
    \caption{LFR-kNN}
\end{subfigure}
\caption{Credit approval: ABROCA slice plots}
\label{fig:credit-approval-abroca}
\end{figure*}

The experimental results on the Credit approval dataset are presented in 
Table \ref{tbl:result-credit-approval} and Figure \ref{fig:credit-approval-abroca}. The dashed lines are used to separate the group of predictive models (traditional, pre-processing, in-processing, and post-processing approaches). The best value is shown in \textbf{bold}, and the second-best value is \underline{underlined}.

It is obvious that (fair) classification models cannot satisfy multiple fairness measures simultaneously. Among fairness-aware models, AdaFair is the best model in terms of accuracy (0.8529) and balanced accuracy (0.8579). Besides, LFR-kNN is a notable model with the best performance on SP, EO, EOd, PE and ABROCA measures. However, its accuracy and balanced accuracy are very low with values 0.5588 and 0.5, respectively. In addition, most pre-processing models prioritize fairness at the cost of significant reductions in accuracy and balanced accuracy. For the TE measures, as defined in Eq. \ref{eq:treatment_equality}, the result may be ``NaN'' if the denominator equals zero. In addition, a limitation of the post-processing models is the inability to obtain ABROCA values, as calculating ABROCA requires model probabilities across multiple thresholds. This is reflected in ``NaN'' values in Table \ref{tbl:result-credit-approval}.
Overall, the in-processing approach outperforms pre-processing and post-processing approaches in terms of trade-off between predictive performance and fairness constraints. 

\begin{table}[h]
\resizebox{\textwidth}{!}
{
\begin{minipage}{\textwidth}
\centering
\caption{Credit approval: performance of predictive models. Protected attribute: Sex}
\label{tbl:result-credit-approval}
\begin{tabular}{lccccccccc}\hline
\textbf{Model} & \textbf{BA} & \textbf{Acc.} & \textbf{SP} & \textbf{EO} & \textbf{EOd} & \textbf{PP} & \textbf{PE} & \textbf{TE} & \textbf{ABROCA} \\ \hline
\textbf{DT} & 0.7646 & 0.7696 & 0.0840 & 0.0962 & 0.1167 & 0.0536 & 0.0205 & 0.2190 & 0.0378 \\
\textbf{NB} & 0.7629 & 0.7794 & 0.0530 & 0.0637 & 0.0729 & 0.0647 & 0.0092 & -0.5357 & 0.0483 \\
\textbf{MLP} & 0.7038 & 0.7107 & 0.0922 & 0.0950 & 0.1432 & 0.0414 & 0.0482 & 0.2697 & 0.1005 \\
\textbf{kNN} & 0.6494 & 0.6617 & -0.1051 & \underline{0.0084} & 0.0638 & 0.1131 & 0.0554 & -0.6150 & 0.0664 \\\hdashline
\textbf{DIR-DT} & 0.5421 & 0.5196 & 0.0137 & 0.1117 & 0.1774 & 0.1241 & 0.0656 & \textbf{0.0142} & 0.0887 \\
\textbf{DIR-NB} & \underline{0.8263} & \underline{0.8333} & -0.0423 & 0.1659 & 0.1874 & 0.0304 & 0.0215 & -1.6500 & 0.0471 \\
\textbf{DIR-MLP} & 0.7360 & 0.7402 & 0.0081 & 0.0649 & 0.1469 & 0.1469 & 0.0820 & -0.3818 & 0.0630 \\
\textbf{DIR-kNN} & 0.6570 & 0.6715 & \underline{0.0058} & 0.0468 & 0.1032 & 0.1348 & 0.0564 & -0.6333 & 0.0668 \\
\textbf{LFR-DT} & 0.5055 & 0.5637 & -0.0154 & 0.0384 & 0.0384 & \textbf{0.0} & \textbf{0.0} & NaN & 0.0897 \\
\textbf{LFR-NB} & 0.5356 & 0.5882 & -0.0184 & 0.0913 & 0.1180 & 0.3333 & 0.0267 & NaN & 0.0487 \\
\textbf{LFR-MLP} & 0.5523 & 0.6030 & -0.0419 & 0.1526 & 0.1793 & 0.2857 & 0.0266 & NaN & 0.0410 \\
\textbf{LFR-kNN} & 0.5 & 0.5588 & \textbf{0.0} & \textbf{0.0} & \textbf{0.0} & \textbf{0.0} & \textbf{0.0} & NaN & \textbf{0.0036} \\\hdashline
\textbf{AdaFair} & \textbf{0.8579} & \textbf{0.8529} & 0.1016 & 0.0216 & 0.1068 & 0.0376 & 0.0851 & 0.2250 & 0.0500 \\
\textbf{Agarwal's} & 0.7851 & 0.7990 & 0.0366 & 0.0180 & \underline{0.0272} & 0.0504 & 0.0092 & -0.7500 & \underline{0.0268} \\\hdashline
\textbf{EOP-DT} & 0.7646 & 0.7696 & 0.0840 & 0.0962 & 0.1167 & 0.0536 & 0.0205 & 0.2190 & NaN \\
\textbf{EOP-NB} & 0.7628 & 0.7794 & 0.0530 & 0.0637 & 0.0729 & 0.0647 & 0.0092 & -0.5357 & NaN \\
\textbf{EOP-MLP} & 0.6938 & 0.7010 & 0.1373 & 0.1334 & 0.2339 & \underline{0.0052} & 0.1005 & 0.7143 & NaN \\
\textbf{EOP-kNN} & 0.6280 & 0.6421 & 0.0203 & 0.1238 & 0.2048 & 0.1948 & 0.0810 & -0.4559 & NaN \\
\textbf{CEP-DT} & 0.7646 & 0.7696 & 0.0840 & 0.0961 & 0.1166 & 0.0536 & 0.0205 & 0.2190 & NaN \\
\textbf{CEP-NB} & 0.7573 & 0.7745 & 0.0458 & 0.0481 & 0.0573 & 0.0616 & 0.0092 & -0.6786 & NaN \\
\textbf{CEP-MLP} & 0.6982 & 0.7059 & 0.1075 & 0.1334 & 0.1816 & 0.0572 & 0.0482 & 0.3947 & NaN \\
\textbf{CEP-kNN} & 0.6304 & 0.6471 & 0.0972 & 0.2007 & 0.2048 & 0.1731 & \underline{0.0041} & \underline{0.1830} & NaN \\
\hline
\end{tabular}
\end{minipage}}
\end{table}

\vspace{-5pt}

\subsubsection{Credit Card Clients Dataset}
\label{subsubsec:result-credit-card}
\begin{figure*}[h]
\centering
\vspace{-5pt}
\begin{subfigure}{.30\linewidth}
    \centering
    \includegraphics[width=\linewidth]{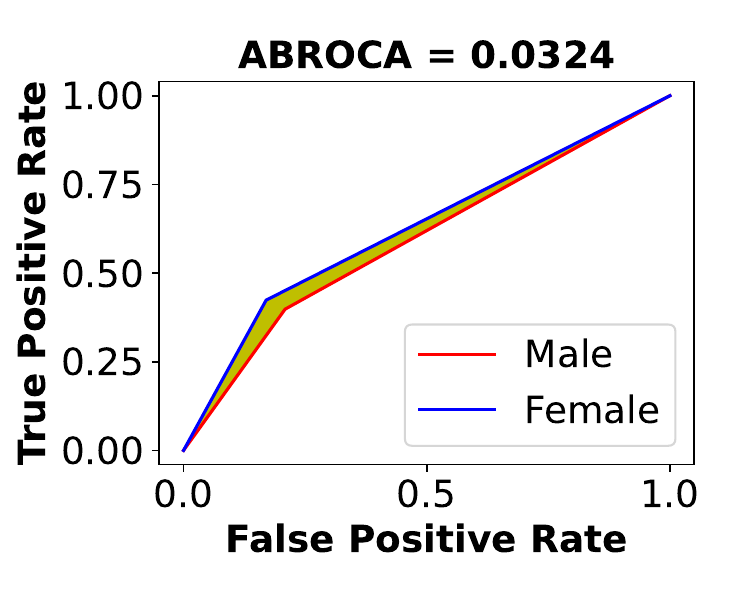}
    \caption{DT}
\end{subfigure}
\hfill
\vspace{-5pt}
\begin{subfigure}{.30\linewidth}
    \centering
    \includegraphics[width=\linewidth]{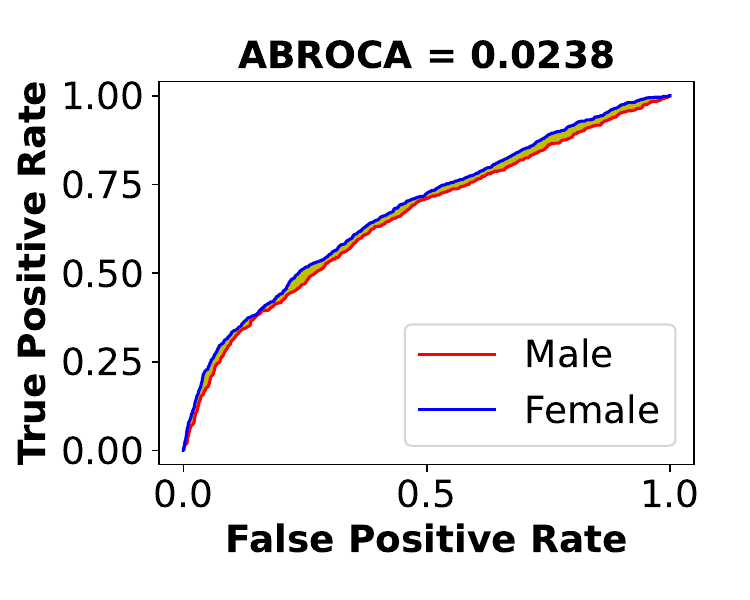}
    \caption{NB}
\end{subfigure}    
\hfill
\vspace{-5pt}
\begin{subfigure}{.30\linewidth}
    \centering
    \includegraphics[width=\linewidth]{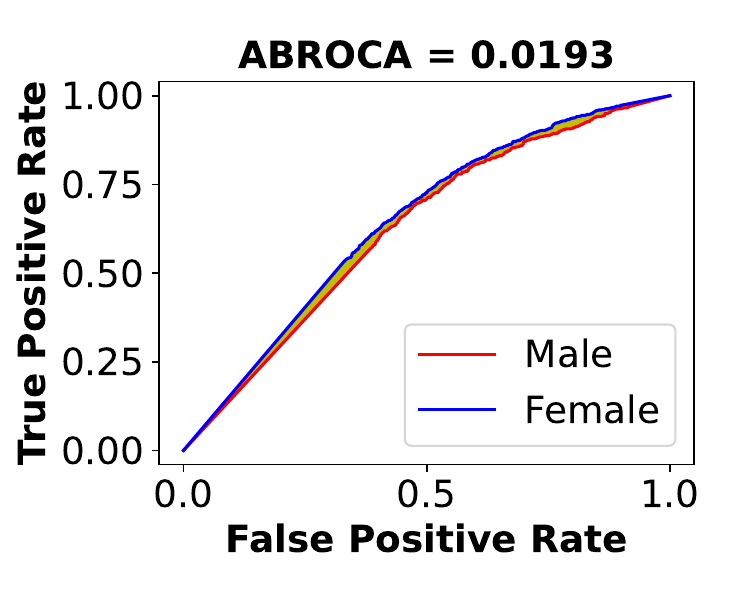}
    \caption{MLP}
\end{subfigure}
\bigskip
\vspace{-5pt}
\begin{subfigure}{.30\linewidth}
    \centering
    \includegraphics[width=\linewidth]{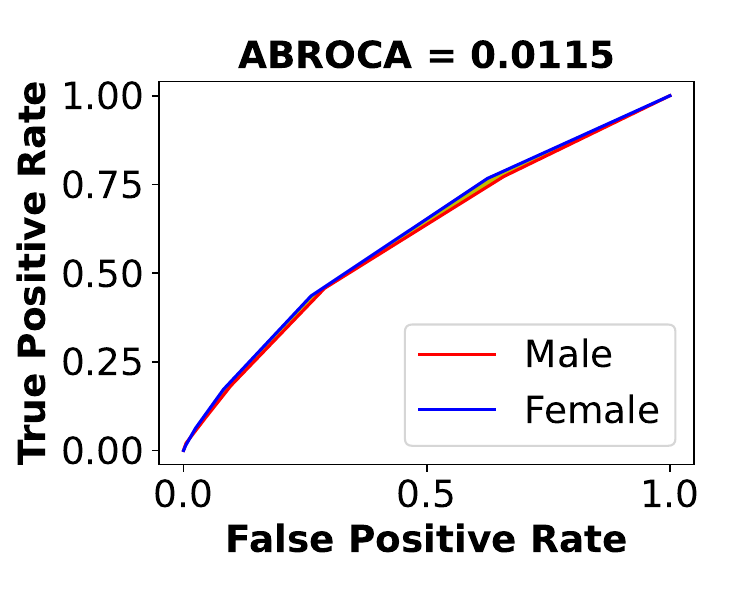}
    \caption{kNN}
\end{subfigure}
\hfill
\vspace{-5pt}
\begin{subfigure}{.30\linewidth}
    \centering
    \includegraphics[width=\linewidth]{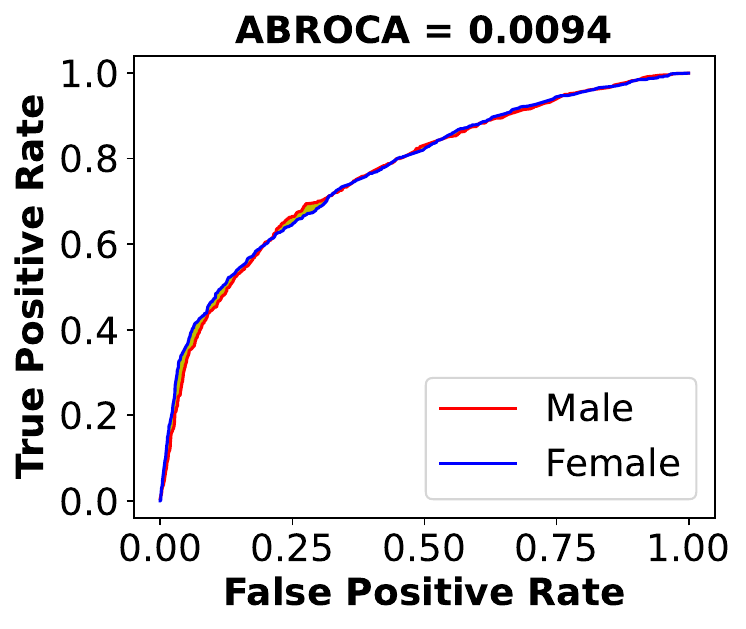}
    \caption{AdaFair}
\end{subfigure}
\vspace{-5pt}
 \hfill
\begin{subfigure}{.30\linewidth}
    \centering
    \includegraphics[width=\linewidth]{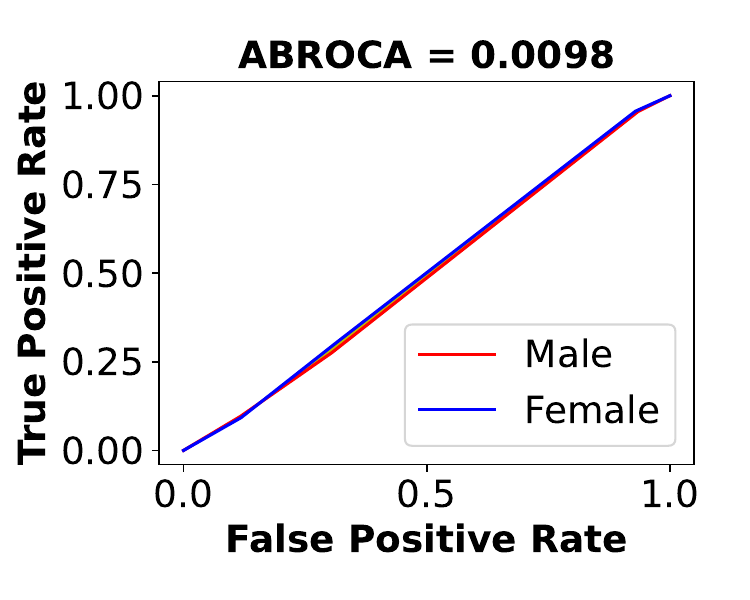}
    \caption{Agarwal's}
\end{subfigure}
\bigskip
\vspace{-5pt}
\begin{subfigure}{.30\linewidth}
    \centering
    \includegraphics[width=\linewidth]{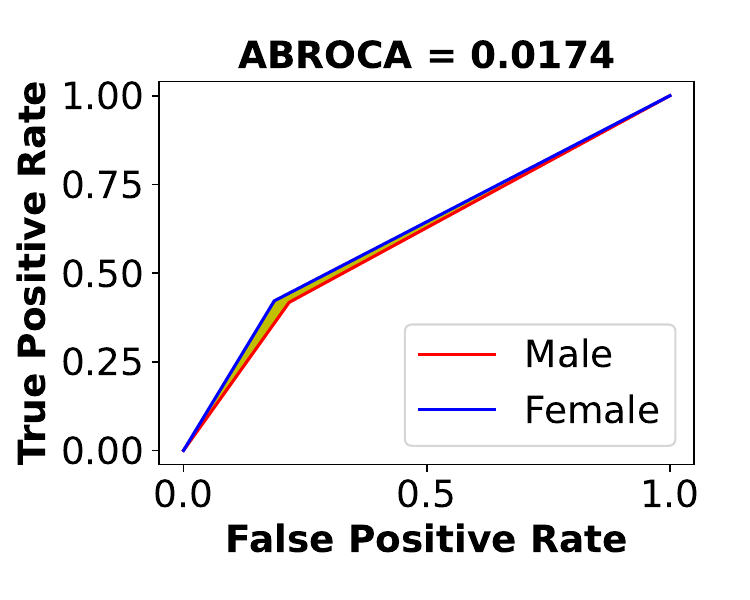}
    \caption{DIR-DT}
\end{subfigure}
\hfill
\vspace{-5pt}
\begin{subfigure}{.30\linewidth}
    \centering
    \includegraphics[width=\linewidth]{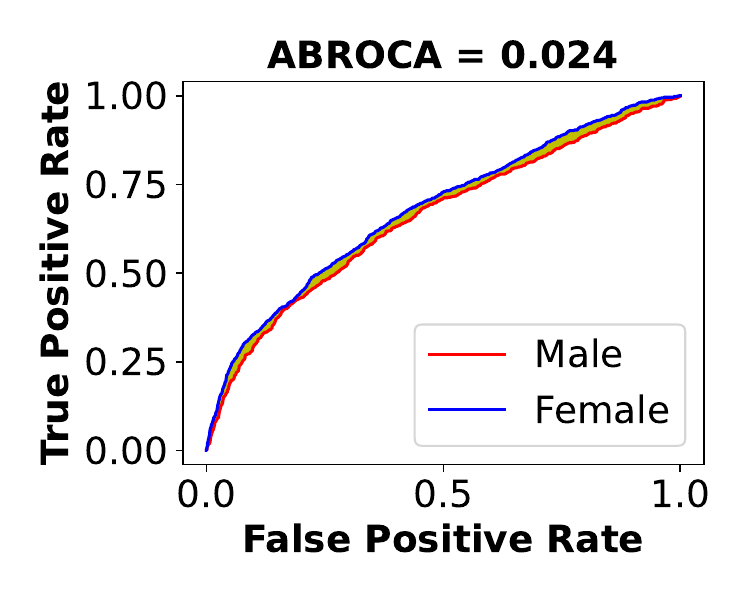}
    \caption{DIR-NB}
\end{subfigure}
\vspace{-5pt}
 \hfill
\begin{subfigure}{.30\linewidth}
    \centering
    \includegraphics[width=\linewidth]{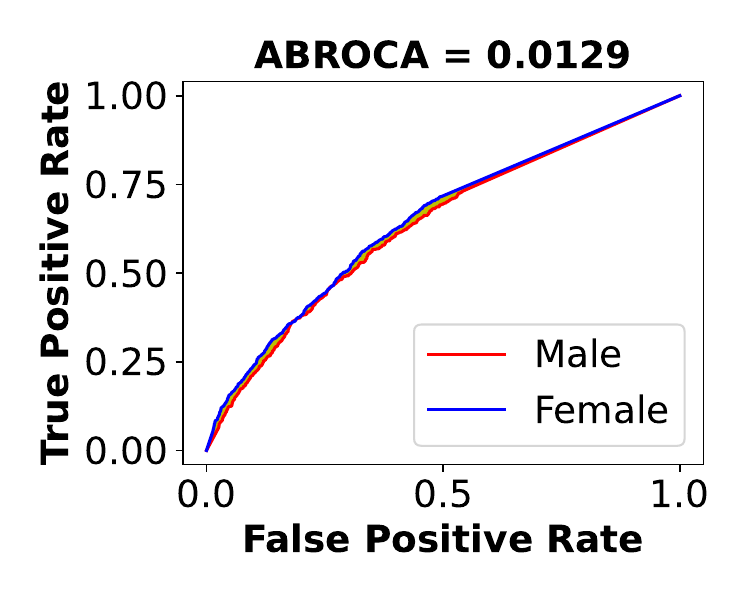}
    \caption{DIR-MLP}
\end{subfigure}
\bigskip
\vspace{-3pt}
\begin{subfigure}{.30\linewidth}
    \centering
    \includegraphics[width=\linewidth]{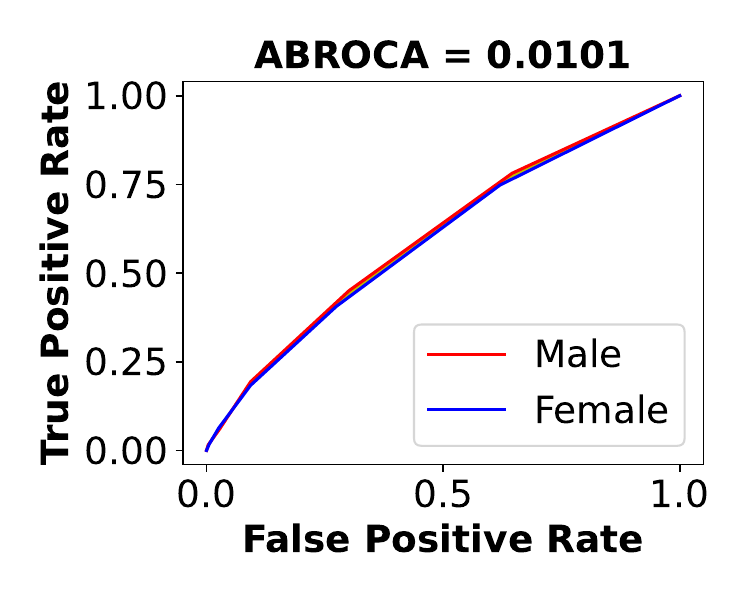}
    \caption{DIR-kNN}
\end{subfigure}
\hfill
\vspace{-3pt}
\begin{subfigure}{.30\linewidth}
    \centering
    \includegraphics[width=\linewidth]{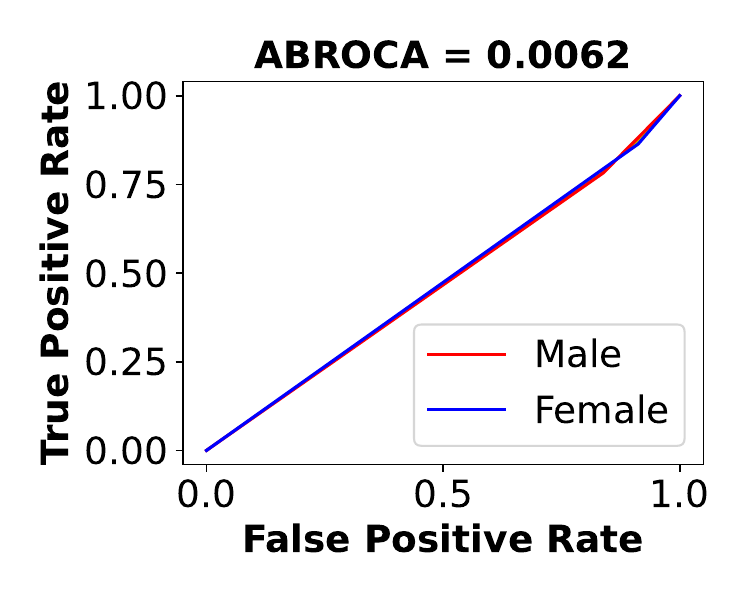}
    \caption{LFR-DT}
\end{subfigure}
\vspace{-3pt}
 \hfill
\begin{subfigure}{.30\linewidth}
    \centering
    \includegraphics[width=\linewidth]{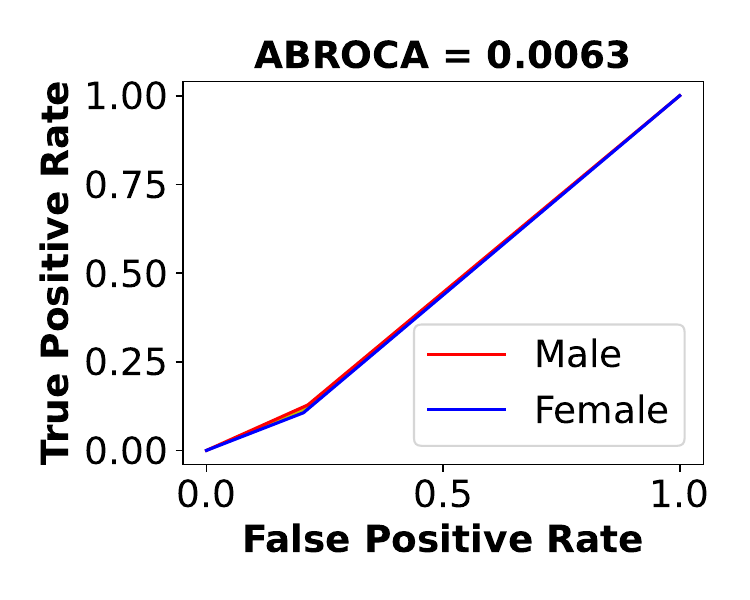}
    \caption{LFR-NB}
\end{subfigure}
\bigskip
\vspace{-5pt}
\begin{subfigure}{.30\linewidth}
    \centering
    \includegraphics[width=\linewidth]{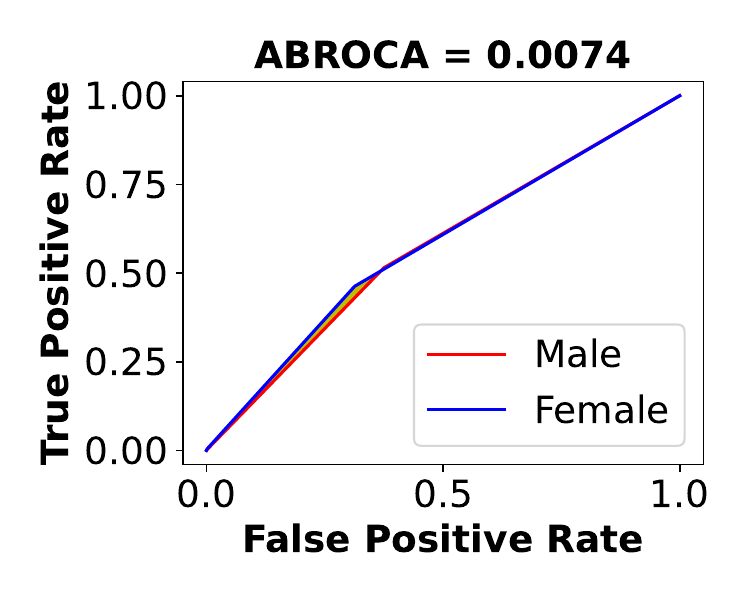}
    \caption{LFR-MLP}
\end{subfigure}
\vspace{-5pt}
\begin{subfigure}{.30\linewidth}
    \centering
    \includegraphics[width=\linewidth]{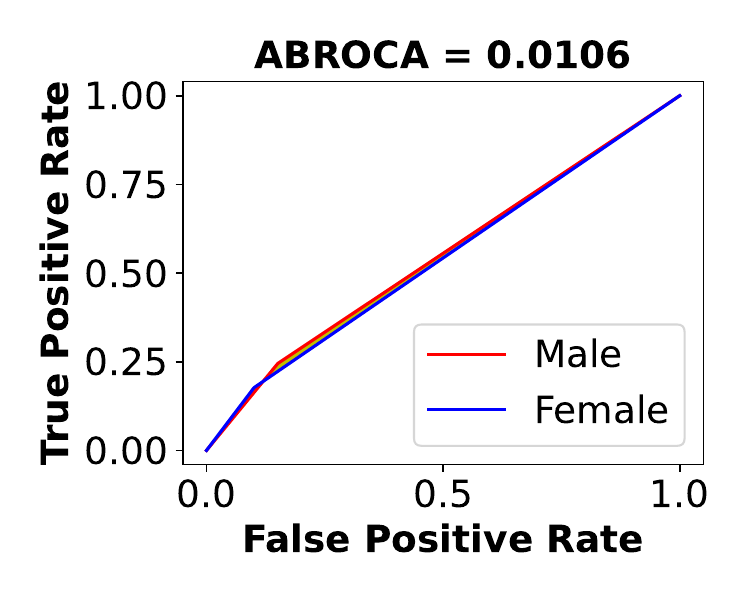}
    \caption{LFR-kNN}
\end{subfigure}
\caption{Credit card: ABROCA slice plots}
\label{fig:credit-card-abroca}
\end{figure*}

Table \ref{tbl:result-credit-card} and Figure \ref{fig:credit-card-abroca} report the experimental results of the predictive models on the credit card clients dataset. AdaFair once again demonstrates its ability to achieve accurate predictions while maintaining fairness, standing out as a fair classification model with the highest accuracy and balanced accuracy, at 0.8160 and 0.6460, respectively. The pre-processing approaches exhibit strong performance on specific fairness measures: LFR-kNN achieves the best results on EOd and TE, DIR-DT performs best on PP, and LFR-DT attains the lowest ABROCA value. Nevertheless, these models generally yield lower accuracy and balanced accuracy compared to the traditional classification models.
\begin{table}[h]
\resizebox{\textwidth}{!}
{
\begin{minipage}{\textwidth}
\centering
\caption{Credit card clients: performance of predictive models. Protected attribute: Sex}
\label{tbl:result-credit-card}

\begin{tabular}{lccccccccc}\hline
\textbf{Model} & \textbf{BA} & \textbf{Acc.} & \textbf{SP} & \textbf{EO} & \textbf{EOd} & \textbf{PP} & \textbf{PE} & \textbf{TE} & \textbf{ABROCA} \\ \hline
\textbf{DT} & 0.6131 & 0.7277 & 0.0308 & 0.0263 & 0.0656 & 0.0275 & 0.0393 & 0.0071 & 0.0324 \\
\textbf{NB} & 0.5599 & 0.3778 & \textbf{-0.0034} & 0.0308 & 0.0311 & 0.0226 &\textbf{0.0002} & -0.0184 & 0.0238 \\
\textbf{MLP} & 0.6111 & 0.5782 & 0.0523 & 0.0403 & 0.0883 & 0.0231 & 0.0481 & 0.0148 & 0.0193 \\
\textbf{kNN} & 0.5435 & 0.7530 & 0.0153 & 0.0089 & 0.0230 & 0.0105 & 0.0142 & 0.0454 & 0.0115 \\\hdashline
\textbf{DIR-DT} & 0.6099 & 0.7187 & 0.0290 & \underline{0.0034} & 0.0342 & \textbf{0.0011} & 0.0308 & -0.0079 & 0.0174 \\
\textbf{DIR-NB} & 0.5674 & 0.4104 & 0.0121 & 0.0174 & 0.0333 & 0.0215 & 0.0158 & -0.0152 & 0.0240 \\
\textbf{DIR-MLP} & 0.5301 & \underline{0.7814} & 0.0245 & 0.0258 & 0.0479 & 0.1022 & 0.0220 & 7.9436 & 0.0129 \\
\textbf{DIR-kNN} & 0.5471 & 0.7511 & 0.0053 & 0.0106 & \underline{0.0111} & 0.0482 & \underline{0.0005} & -0.3479 & 0.0101 \\
\textbf{LFR-DT} & 0.5798 & 0.5897 & 0.0476 & 0.0039 & 0.0586 & 0.0061 & 0.0546 & \underline{-0.0043} & \textbf{0.0062} \\
\textbf{LFR-NB} & 0.4831 & 0.7103 & -0.0053 & 0.0081 & 0.0161 & 0.0457 & 0.0079 & -0.5013 & \underline{0.0063} \\
\textbf{LFR-MLP} & 0.4514 & 0.6406 & 0.0117 & 0.0264 & 0.0373 & 0.0384 & 0.0109 & -0.8892 & 0.0074 \\
\textbf{LFR-kNN} & 0.4967 & 0.2270 & -0.0044 & 0.0051 & \textbf{0.0091} & 0.0267 & 0.0040 & \textbf{-0.0026} & 0.0106 \\\hdashline
\textbf{AdaFair} & \textbf{0.6460} & \textbf{0.8160} & 0.0062 & 0.0321 & 0.0391 & 0.0155 & 0.0070 & -0.2601 & 0.0094 \\
\textbf{Agarwal's} & 0.5025 & 0.5270 & 0.0045 & 0.0228 & 0.0238 & 0.0362 & 0.0009 & -0.0392 & 0.0098 \\\hdashline
\textbf{EOP-DT} & \underline{0.6132} & 0.7278 & 0.0305 & 0.0263 & 0.0652 & 0.0271 & 0.0389 & 0.0056 & NaN \\
\textbf{EOP-NB} & 0.5548 & 0.3714 & 0.0090 & 0.0379 & 0.0566 & 0.0165 & 0.0187 & -0.0199 & NaN \\
\textbf{EOP-MLP} & 0.6073 & 0.5812 & 0.0138 & \textbf{0.0026} & 0.0137 & 0.0267 & 0.0110 & -0.0311 & NaN \\
\textbf{EOP-kNN} & 0.5416 & 0.7534 & 0.0062 & 0.0053 & 0.0119 & 0.0107 & 0.0066 & -0.2388 & NaN \\
\textbf{CEP-DT} & 0.6131 & 0.7277 & 0.0308 & 0.0263 & 0.0656 & 0.0275 & 0.0393 & 0.0071 & NaN \\
\textbf{CEP-NB} & 0.5599 & 0.3778 & \underline{-0.0035} & 0.0308 & 0.0311 & 0.0226 & \textbf{0.0002} & -0.0184 & NaN \\
\textbf{CEP-MLP} & 0.6111 & 0.5782 & 0.0522 & 0.0403 & 0.0883 & 0.0231 & 0.0481 & 0.0148 & NaN \\
\textbf{CEP-kNN} & 0.5407 & 0.7561 & 0.0302 & 0.0322 & 0.0590 & \underline{0.0051} & 0.0268 & 0.6475 & NaN \\
\hline

\end{tabular}
\end{minipage}}
\end{table}

\subsubsection{Credit Scoring Dataset}
\label{subsubsec:result-credit-scoring}
\begin{figure*}[h]
\centering
\begin{subfigure}{.30\linewidth}
    \centering
    \includegraphics[width=\linewidth]{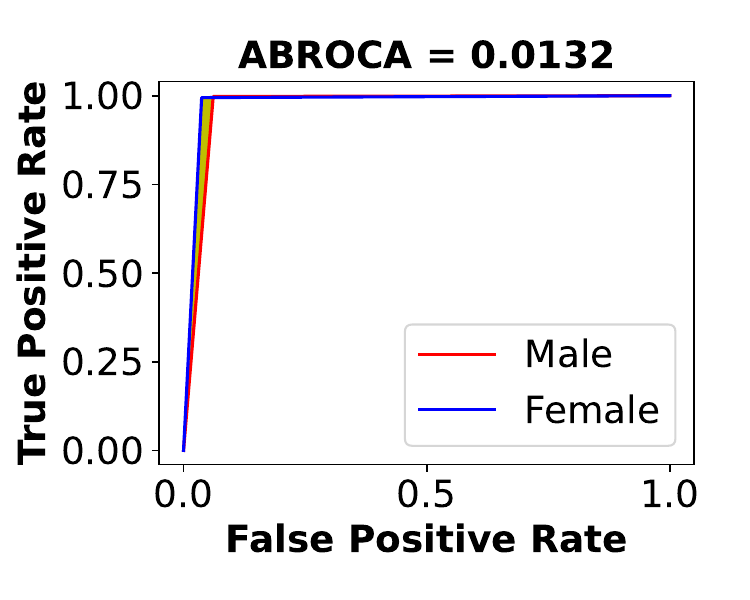}
    \caption{DT}
\end{subfigure}
\hfill
\vspace{-5pt}
\begin{subfigure}{.30\linewidth}
    \centering
    \includegraphics[width=\linewidth]{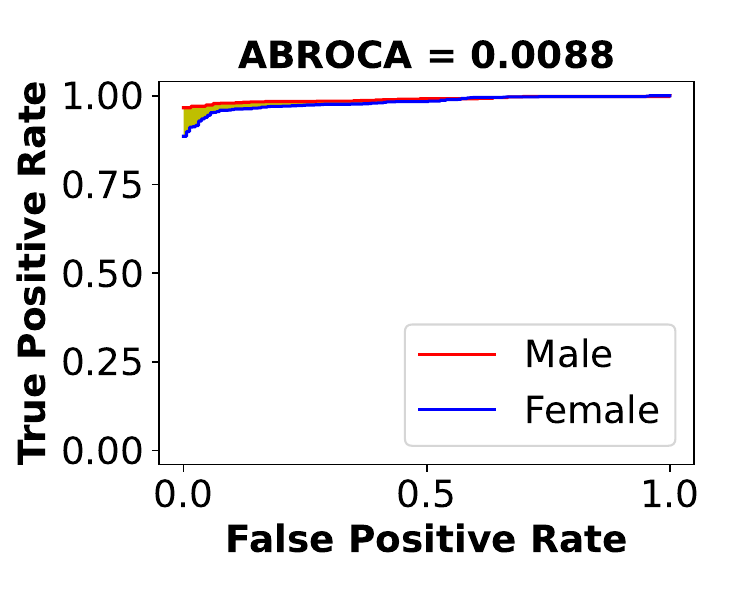}
    \caption{NB}
\end{subfigure}    
\hfill
\vspace{-5pt}
\begin{subfigure}{.30\linewidth}
    \centering
    \includegraphics[width=\linewidth]{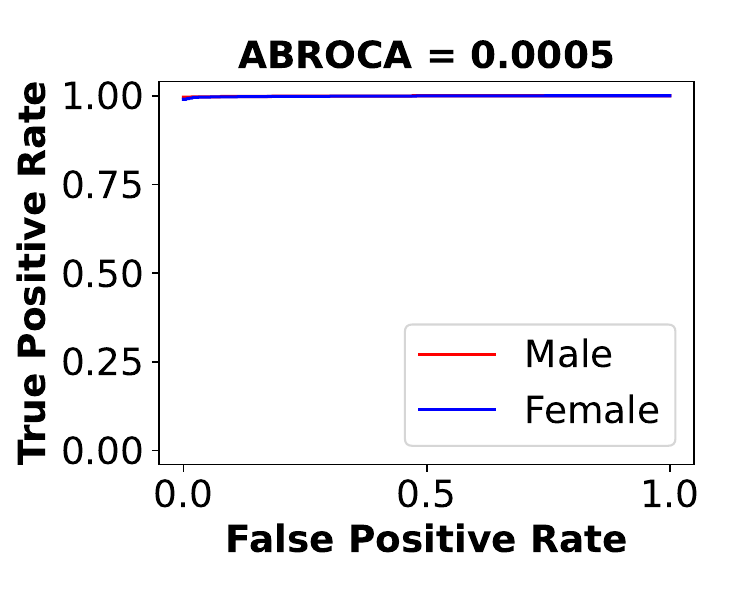}
    \caption{MLP}
\end{subfigure}
\bigskip
\vspace{-5pt}
\begin{subfigure}{.30\linewidth}
    \centering
    \includegraphics[width=\linewidth]{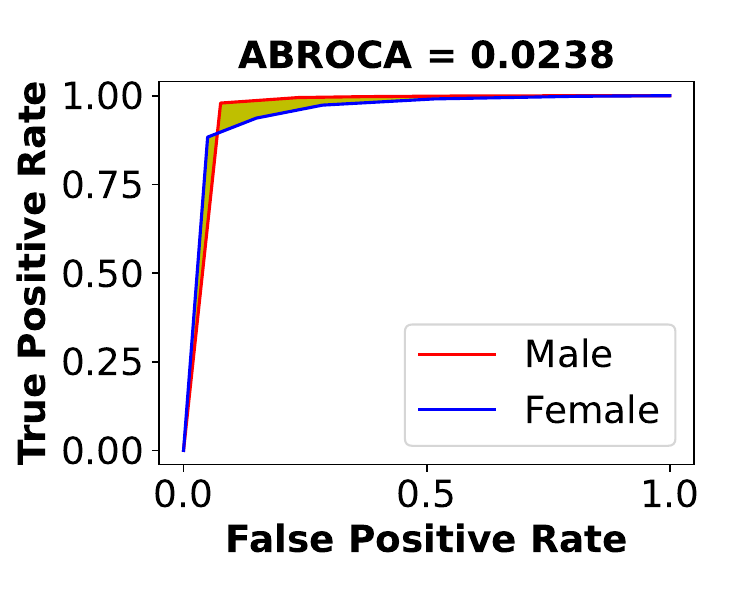}
    \caption{kNN}
\end{subfigure}
\hfill
\vspace{-5pt}
\begin{subfigure}{.30\linewidth}
    \centering
    \includegraphics[width=\linewidth]{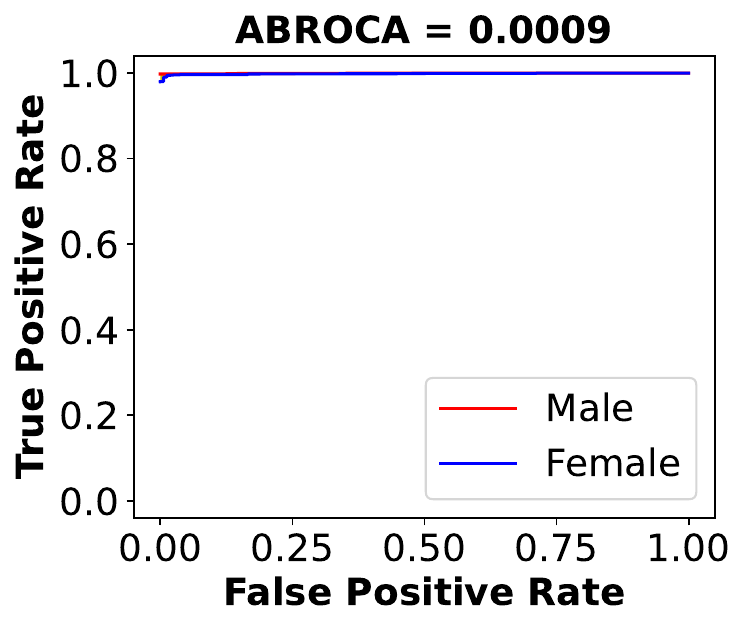}
    \caption{AdaFair}
\end{subfigure}
\vspace{-5pt}
 \hfill
\begin{subfigure}{.30\linewidth}
    \centering
    \includegraphics[width=\linewidth]{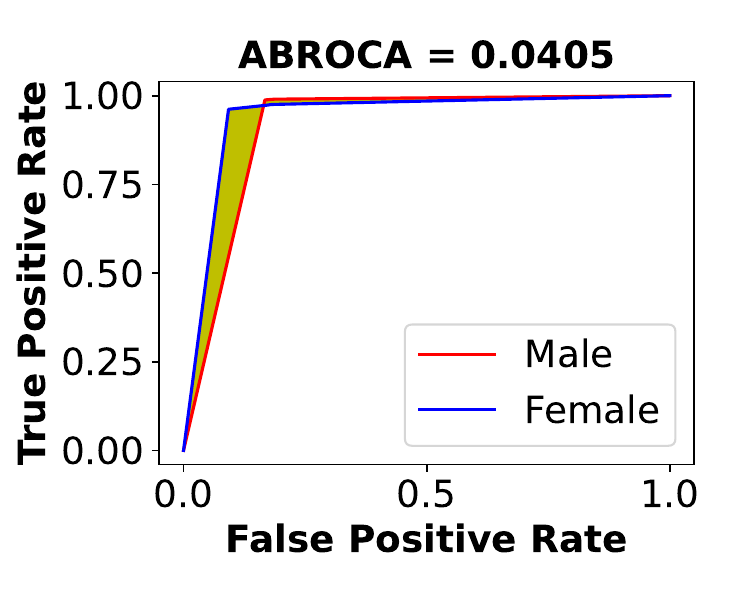}
    \caption{Agarwal's}
\end{subfigure}
\bigskip
\vspace{-5pt}
\begin{subfigure}{.30\linewidth}
    \centering
    \includegraphics[width=\linewidth]{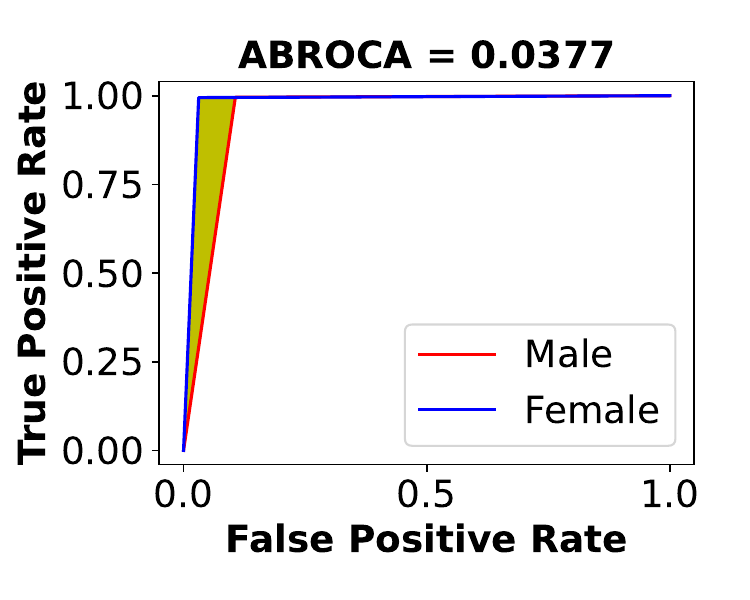}
    \caption{DIR-DT}
\end{subfigure}
\hfill
\vspace{-5pt}
\begin{subfigure}{.30\linewidth}
    \centering
    \includegraphics[width=\linewidth]{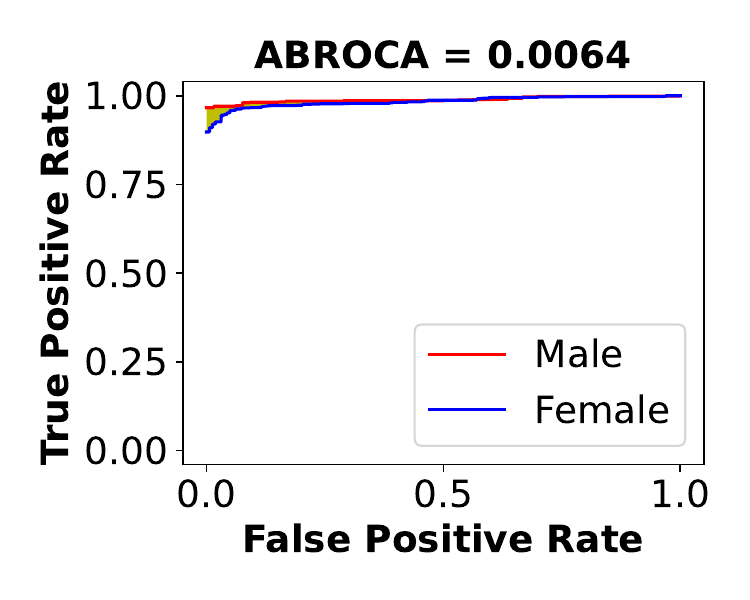}
    \caption{DIR-NB}
\end{subfigure}
\vspace{-5pt}
 \hfill
\begin{subfigure}{.30\linewidth}
    \centering
    \includegraphics[width=\linewidth]{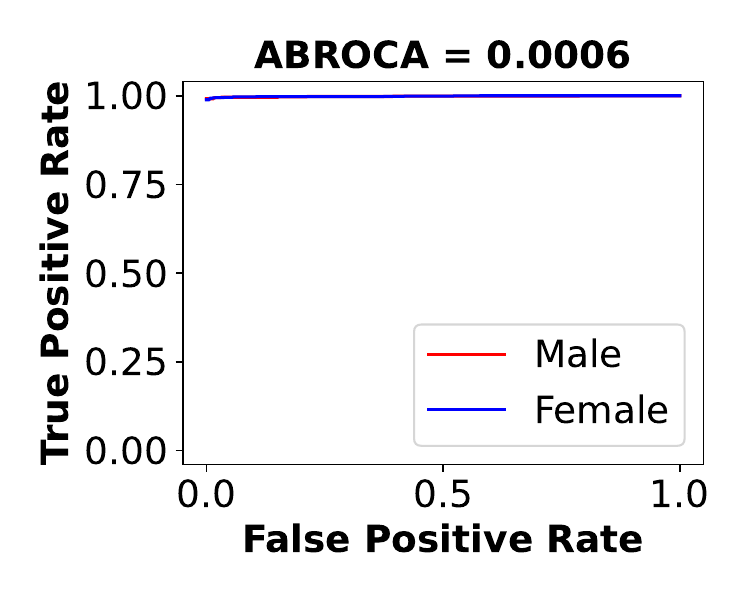}
    \caption{DIR-MLP}
\end{subfigure}
\bigskip
\vspace{-5pt}
\begin{subfigure}{.30\linewidth}
    \centering
    \includegraphics[width=\linewidth]{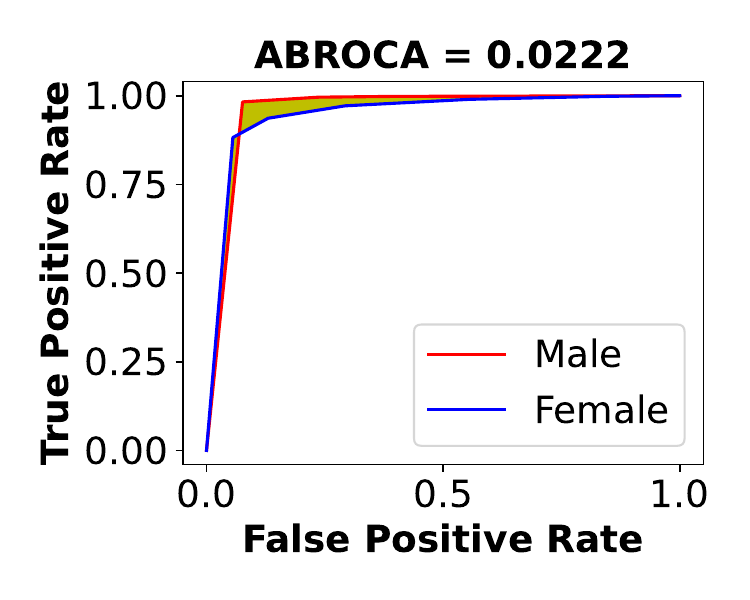}
    \caption{DIR-kNN}
\end{subfigure}
\hfill
\vspace{-5pt}
\begin{subfigure}{.30\linewidth}
    \centering
    \includegraphics[width=\linewidth]{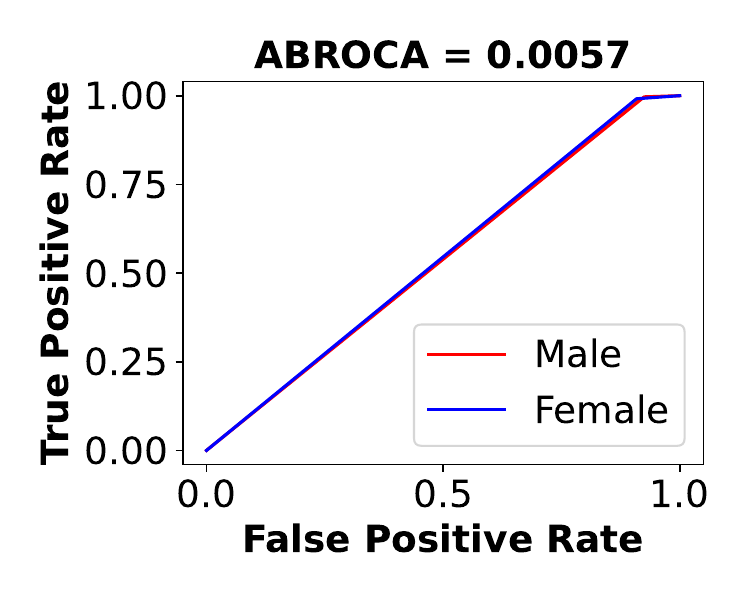}
    \caption{LFR-DT}
\end{subfigure}
\vspace{-5pt}
 \hfill
\begin{subfigure}{.30\linewidth}
    \centering
    \includegraphics[width=\linewidth]{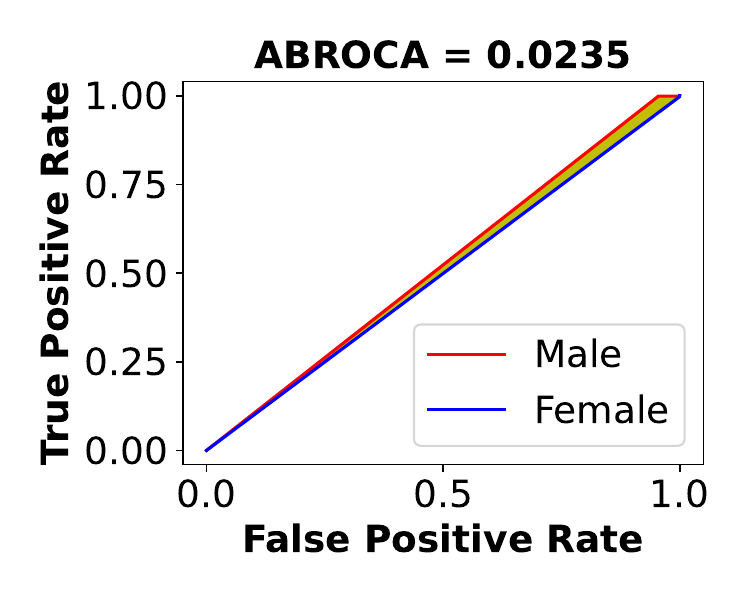}
    \caption{LFR-NB}
\end{subfigure}
\bigskip
\vspace{-5pt}
\begin{subfigure}{.30\linewidth}
    \centering
    \includegraphics[width=\linewidth]{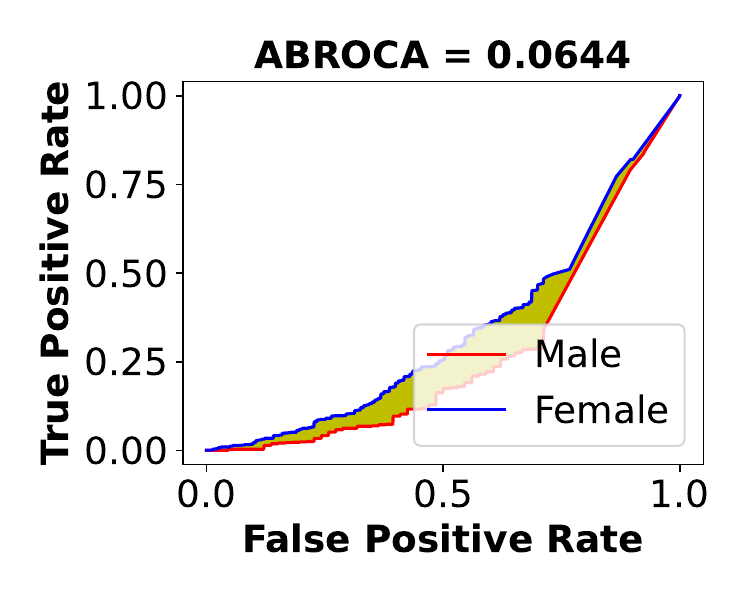}
    \caption{LFR-MLP}
\end{subfigure}
\vspace{-5pt}
\begin{subfigure}{.30\linewidth}
    \centering
    \includegraphics[width=\linewidth]{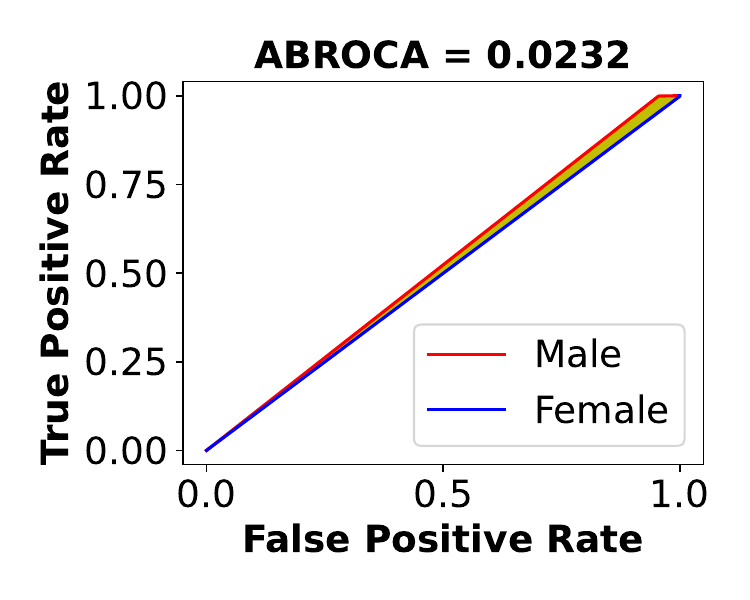}
    \caption{LFR-kNN}
\end{subfigure}
\caption{Credit scoring: ABROCA slice plots}
\label{fig:credit-scoring-abroca}
\end{figure*}

The experimental results on the credit scoring dataset are presented in Table \ref{tbl:result-credit-scoring} and Figure \ref{fig:credit-scoring-abroca}. In terms of predictive performance, AdaFair once again outperforms the other models, achieving very high accuracy (0.9943) and balanced accuracy (0.9929). In addition, the pre-processing model LFR-MLP attains perfect scores on multiple fairness measures, including SP, EO, EOd, PE, and TE; however, its balanced accuracy remains low at 0.5. The traditional MLP model stands out by achieving the best performance on the ABROCA fairness measure (0.0005). By contrast, the pre-processing LFR model performs poorly due to its very low balanced accuracy (below 0.5), whereas the DIR model demonstrates a more favorable balance by achieving both accurate classification performance and strong fairness outcomes.

\begin{table}[h]
\resizebox{\textwidth}{!}
{
\begin{minipage}{\textwidth}
\centering
\caption{Credit scoring: performance of predictive models. Protected attribute: Sex}
\label{tbl:result-credit-scoring}

\begin{tabular}{lccccccccc}\hline
\textbf{Model} & \textbf{BA} & \textbf{Acc.} & \textbf{SP} & \textbf{EO} & \textbf{EOd} & \textbf{PP} & \textbf{PE} & \textbf{TE} & \textbf{ABROCA} \\ \hline
\textbf{DT} & 0.9761 & 0.9924 & 0.0345 & 0.0032 & 0.0269 & 0.0003 & 0.0238 & 0.8333 & 0.0132 \\
\textbf{NB} & 0.8785 & 0.9585 & 0.0448 & 0.0149 & 0.1126 & 0.0003 & 0.0976 & 0.7148 & 0.0088 \\
\textbf{MLP} & \underline{0.9923} & 0.9931 & 0.0299 & \underline{0.0001} & 0.0124 & 0.0013 & 0.0122 & NaN & \textbf{0.0005} \\
\textbf{kNN} & 0.8447 & 0.9581 & 0.0460 & 0.0236 & 0.0596 & 0.0082 & 0.0359 & 0.7267 & 0.0238 \\\hdashline
\textbf{DIR-DT} & 0.9713 & 0.9908 & 0.0361 & 0.0009 & 0.0764 & 0.0043 & 0.0754 & 1.0286 & 0.0377 \\
\textbf{DIR-NB} & 0.9612 & 0.9292 & 0.0934 & 0.0702 & 0.0702 & \textbf{0.0} & \textbf{0.0} & NaN & 0.0064 \\
\textbf{DIR-MLP} & 0.9820 & 0.9924 & 0.0284 & \underline{0.0001} & 0.0218 & 0.0029 & 0.0217 & -3.6667 & \underline{0.0006} \\
\textbf{DIR-kNN} & 0.8378 & 0.9562 & 0.0503 & 0.0267 & 0.0868 & 0.0068 & 0.0601 & 0.8279 & 0.0222 \\
\textbf{LFR-DT} & 0.4674 & 0.7632 & \underline{0.0143} & 0.0191 & 0.0360 & 0.0365 & 0.0169 & -0.7024 & 0.0057 \\
\textbf{LFR-NB} & 0.4683 & 0.8549 & -0.0611 & 0.0642 & 0.0642 & 0.0276 & \textbf{0.0} & -1.0684 & 0.0235 \\
\textbf{LFR-MLP} & 0.5 & 0.9128 & \textbf{0.0} & \textbf{0.0} & \textbf{0.0} & 0.0312 & \textbf{0.0} & \textbf{0.0} & 0.0644 \\
\textbf{LFR-kNN} & 0.4674 & 0.7632 & \underline{0.0143} & 0.0191 & 0.0360 & 0.0365 & 0.0169 & -0.7024 & 0.0232 \\\hdashline
\textbf{AdaFair} & \textbf{0.9929} & \textbf{0.9943} & 0.0298 & \underline{0.0001} & 0.0124 & 0.0014 & 0.0123 & NaN & 0.0009 \\
\textbf{Agarwal's} & 0.9077 & 0.9649 & 0.0421 & 0.0157 & 0.0442 & 0.0038 & 0.0284 & 0.7200 & 0.0405 \\\hdashline
\textbf{EOP-DT} & 0.9761 & 0.9924 & 0.0345 & 0.0032 & 0.0269 & 0.0030 & 0.0238 & 0.8333 & NaN \\
\textbf{EOP-NB} & 0.8598 & 0.9532 & 0.0273 & 0.0012 & 0.0498 & 0.0053 & 0.0486 & -0.2118 & NaN \\
\textbf{EOP-MLP} & 0.9901 & \underline{0.9928} & 0.0293 & \underline{0.0001} & 0.0185 & 0.0020 & 0.0184 & NaN & NaN \\
\textbf{EOP-kNN} & 0.8317 & 0.9524 & 0.0311 & 0.0126 & 0.0286 & 0.0129 & 0.0160 & \underline{0.1192} & NaN \\
\textbf{CEP-DT} & 0.9761 & 0.9924 & 0.0345 & 0.0032 & 0.0270 & 0.0034 & 0.0238 & 0.8333 & NaN \\
\textbf{CEP-NB} & 0.8617 & 0.9566 & 0.0561 & 0.0182 & 0.2371 & 0.0081 & 0.2189 & 1.0442 & NaN \\
\textbf{CEP-MLP} & 0.9901 & \underline{0.9928} & 0.0309 & \underline{0.0001} & \underline{0.0030} & \underline{0.0002} & \underline{0.0029} & -1.000 & NaN \\
\textbf{CEP-kNN} & 0.8229 & 0.9543 & 0.0563 & 0.0236 & 0.2111 & 0.0022 & 0.1875 & 0.7728 & NaN \\
\hline
\end{tabular}
\end{minipage}}
\end{table}

\vspace{-5pt}

\subsubsection{German Credit Dataset}
\label{subsubsec:result-german-credit}
\begin{figure*}[h]
\centering
\begin{subfigure}{.30\linewidth}
    \centering
    \includegraphics[width=\linewidth]{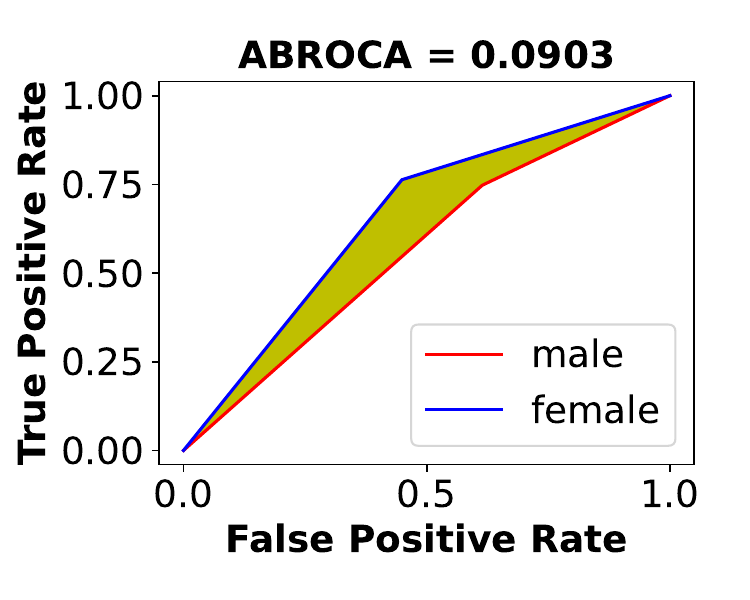}
    \caption{DT}
\end{subfigure}
\hfill
\vspace{-5pt}
\begin{subfigure}{.30\linewidth}
    \centering
    \includegraphics[width=\linewidth]{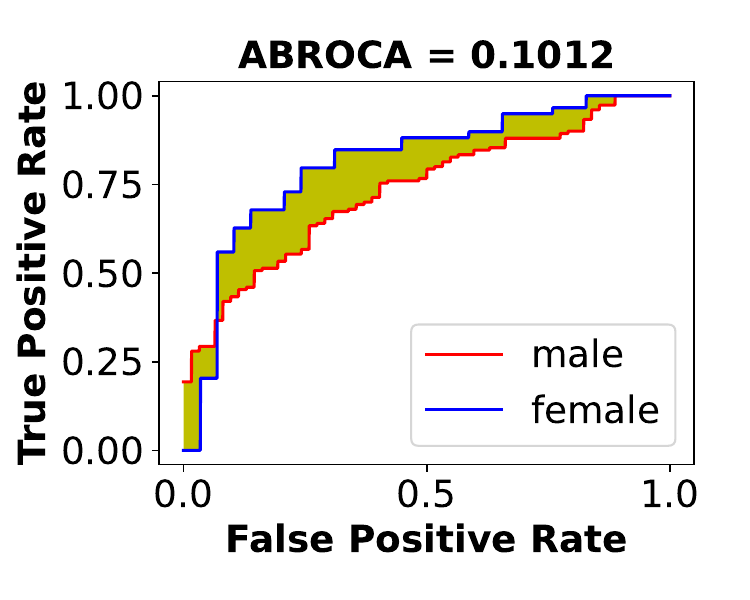}
    \caption{NB}
\end{subfigure}    
\hfill
\vspace{-5pt}
\begin{subfigure}{.30\linewidth}
    \centering
    \includegraphics[width=\linewidth]{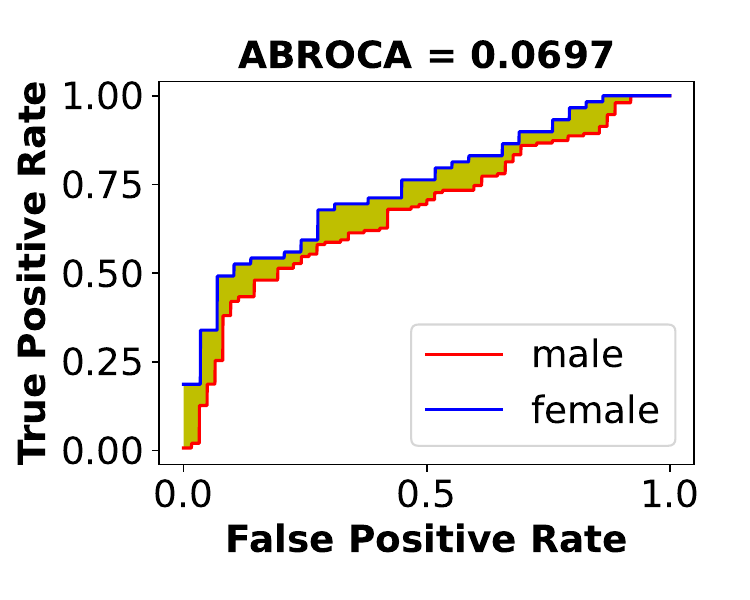}
    \caption{MLP}
\end{subfigure}
\bigskip
\vspace{-5pt}
\begin{subfigure}{.30\linewidth}
    \centering
    \includegraphics[width=\linewidth]{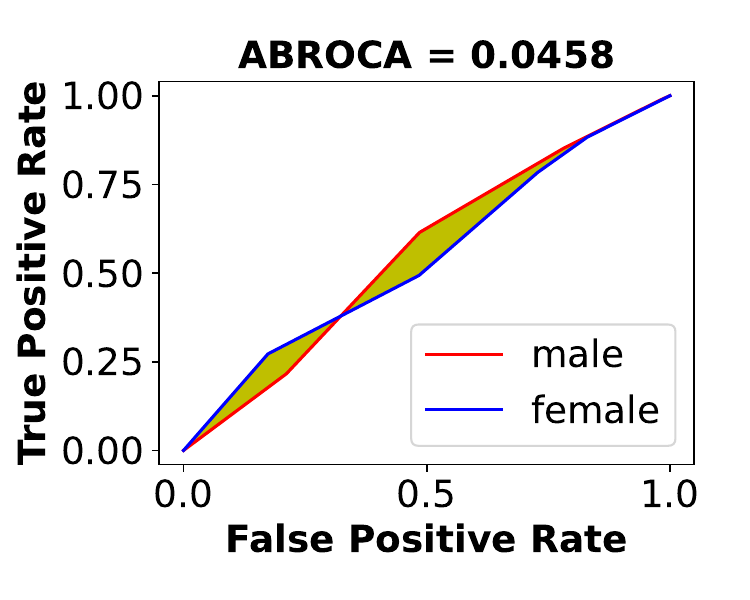}
    \caption{kNN}
\end{subfigure}
\hfill
\vspace{-5pt}
\begin{subfigure}{.30\linewidth}
    \centering
    \includegraphics[width=\linewidth]{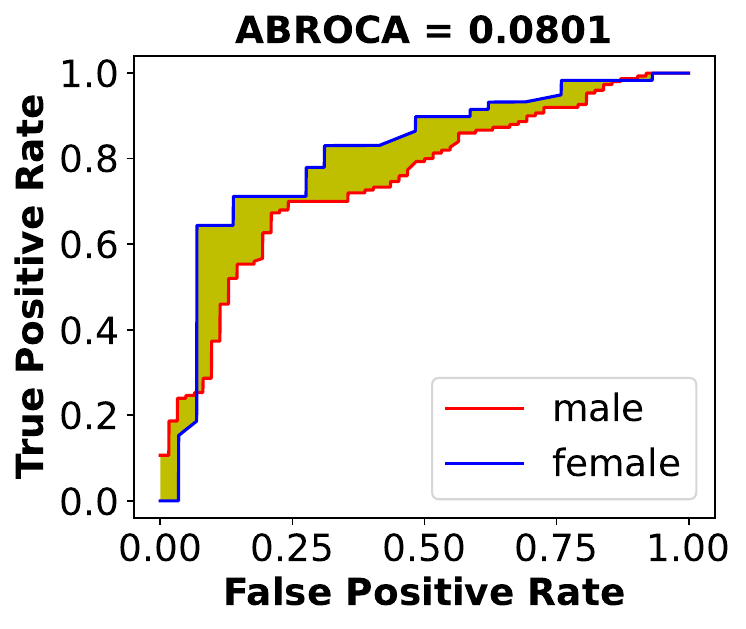}
    \caption{AdaFair}
\end{subfigure}
\vspace{-5pt}
 \hfill
\begin{subfigure}{.30\linewidth}
    \centering
    \includegraphics[width=\linewidth]{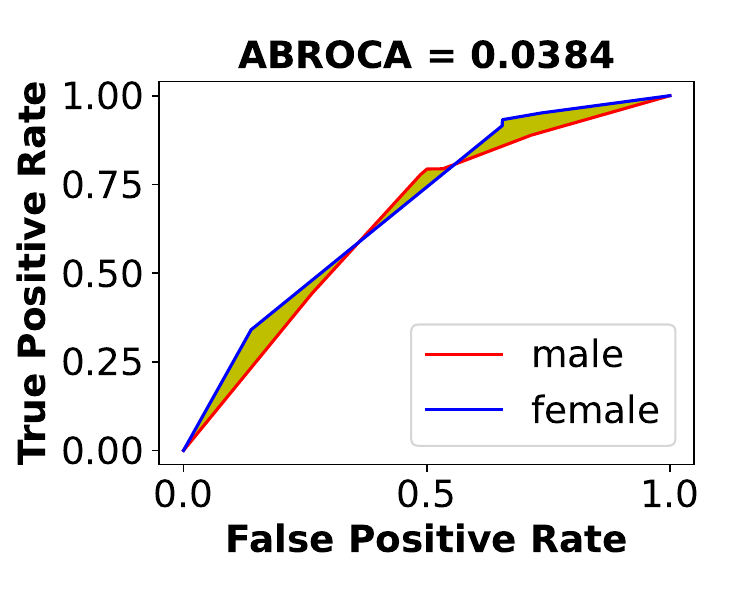}
    \caption{Agarwal's}
\end{subfigure}
\bigskip
\vspace{-5pt}
\begin{subfigure}{.30\linewidth}
    \centering
    \includegraphics[width=\linewidth]{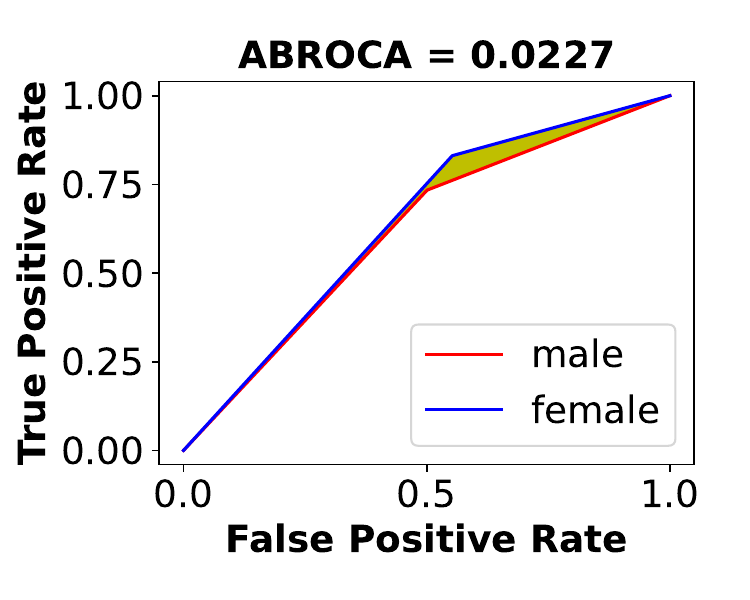}
    \caption{DIR-DT}
\end{subfigure}
\hfill
\vspace{-5pt}
\begin{subfigure}{.30\linewidth}
    \centering
    \includegraphics[width=\linewidth]{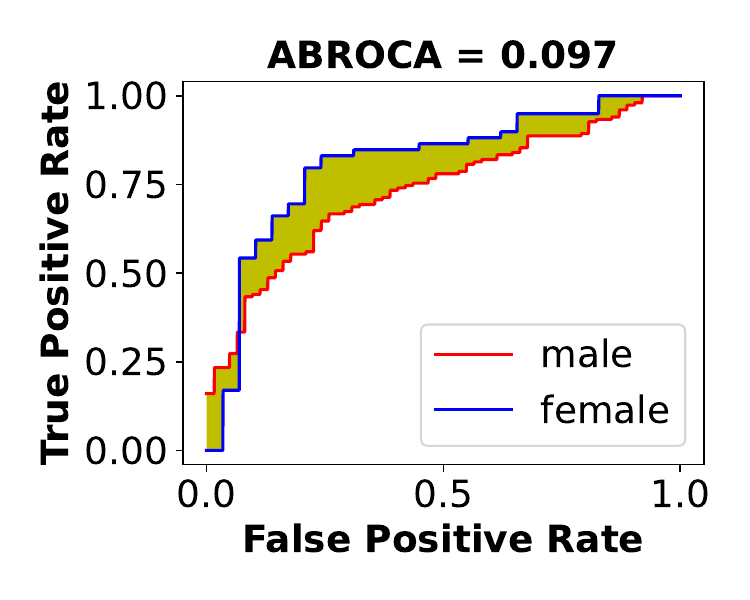}
    \caption{DIR-NB}
\end{subfigure}
\vspace{-5pt}
 \hfill
\begin{subfigure}{.30\linewidth}
    \centering
    \includegraphics[width=\linewidth]{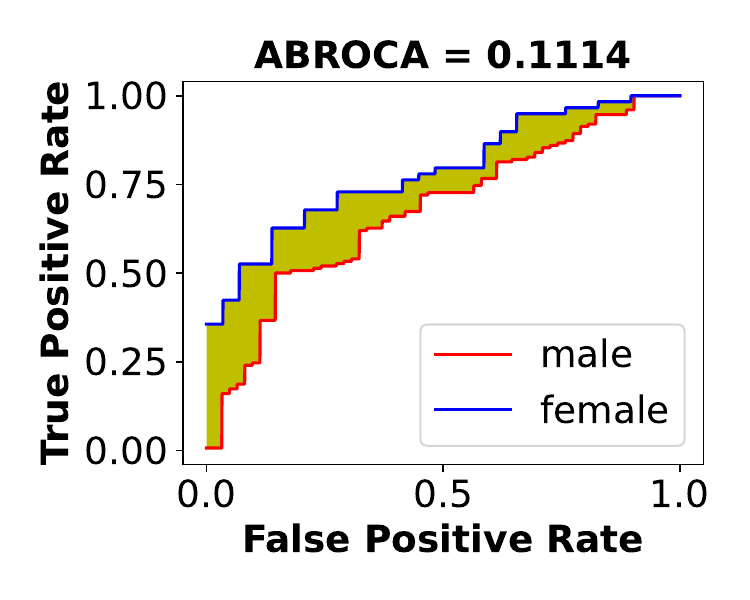}
    \caption{DIR-MLP}
\end{subfigure}
\bigskip
\vspace{-5pt}
\begin{subfigure}{.30\linewidth}
    \centering
    \includegraphics[width=\linewidth]{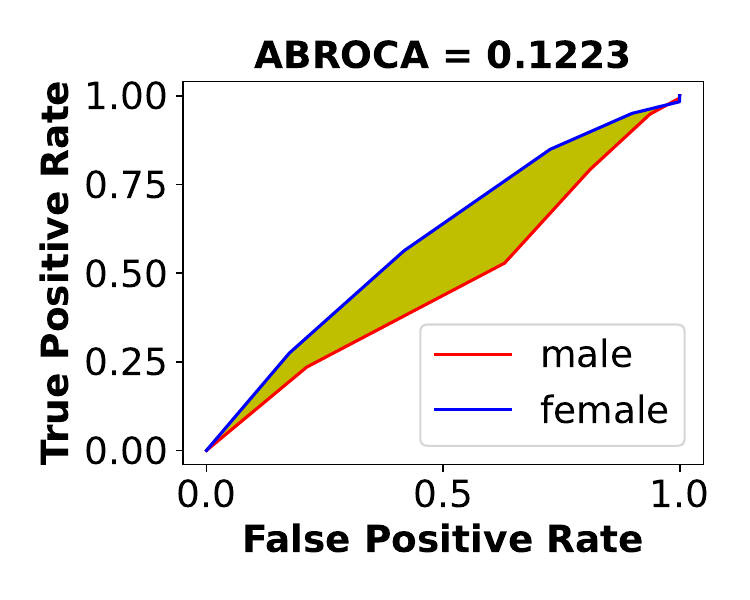}
    \caption{DIR-kNN}
\end{subfigure}
\hfill
\vspace{-5pt}
\begin{subfigure}{.30\linewidth}
    \centering
    \includegraphics[width=\linewidth]{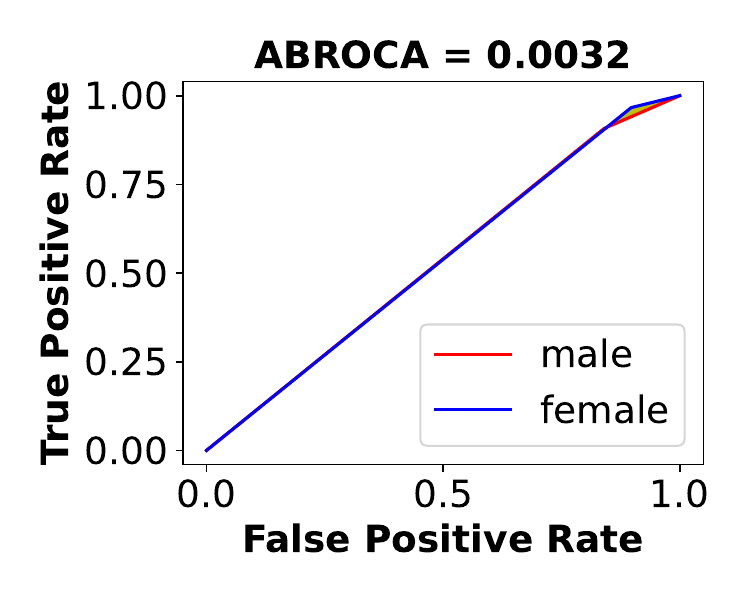}
    \caption{LFR-DT}
\end{subfigure}
\vspace{-5pt}
 \hfill
\begin{subfigure}{.30\linewidth}
    \centering
    \includegraphics[width=\linewidth]{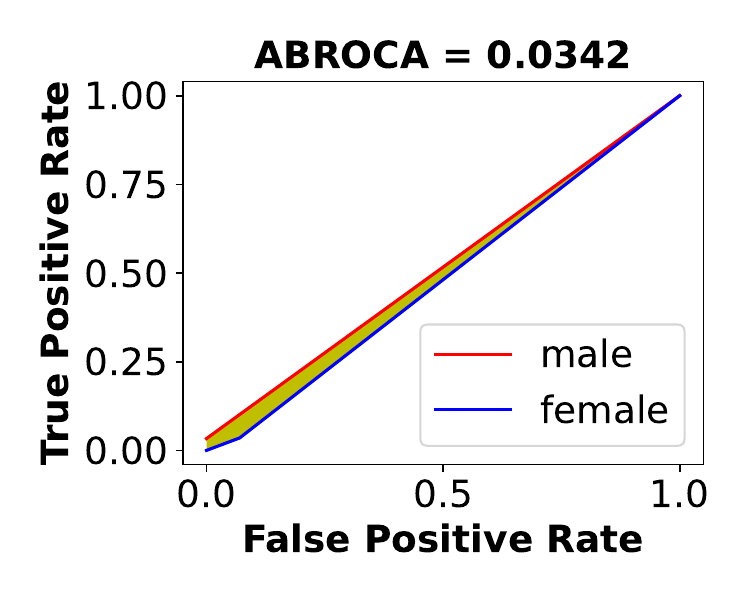}
    \caption{LFR-NB}
\end{subfigure}
\bigskip
\vspace{-5pt}
\begin{subfigure}{.30\linewidth}
    \centering
    \includegraphics[width=\linewidth]{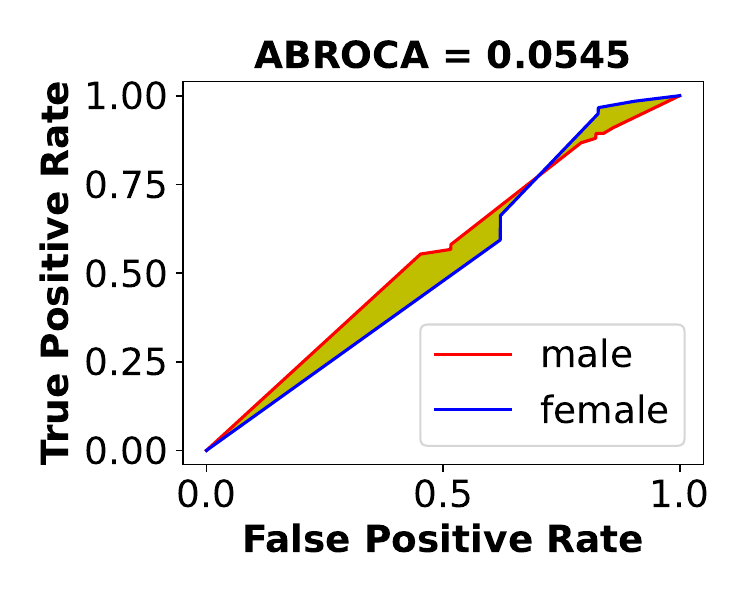}
    \caption{LFR-MLP}
\end{subfigure}
\vspace{-5pt}
\begin{subfigure}{.30\linewidth}
    \centering
    \includegraphics[width=\linewidth]{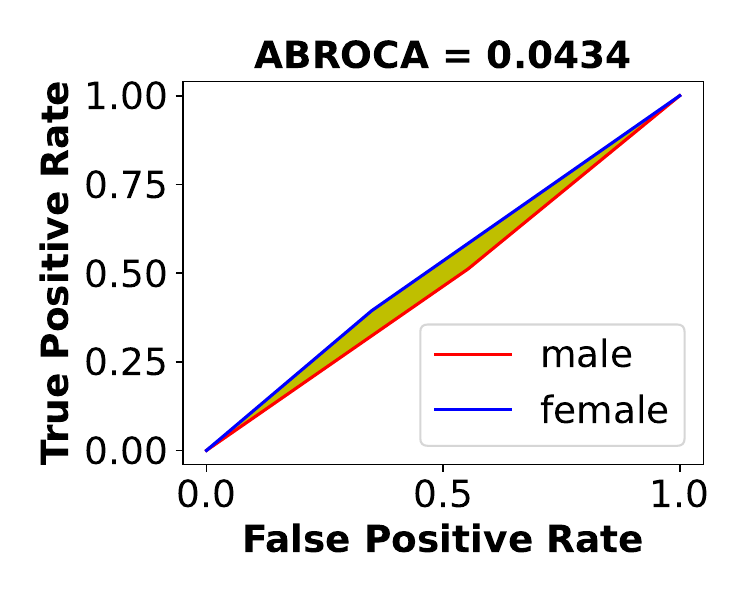}
    \caption{LFR-kNN}
\end{subfigure}
\caption{German credit: ABROCA slice plots}
\label{fig:german-credit-abroca}
\end{figure*}

Table \ref{tbl:result-german-credit} and Figure \ref{fig:german-credit-abroca} present the experimental results of the predictive models on the German Credit dataset. In terms of classification performance, the traditional NB model achieves the best results, with an accuracy of 0.7300 and a balanced accuracy of 0.6604. With respect to fairness constraints, LFR-MLP emerges as the top-performing model, achieving perfect scores on multiple fairness measures, including SP, EO, EOd, PE, and TE. However, its balanced accuracy remains low at 0.5. Among the pre-processing approaches, DIR proves effective in achieving a more favorable trade-off between predictive performance and fairness. Although the post-processing models produce relatively fair classification outcomes, they do not demonstrate strong performance with respect to the fairness constraints.
\begin{table}[h]
\resizebox{\textwidth}{!}
{
\begin{minipage}{\textwidth}
\centering
\caption{German credit: performance of predictive models. Protected attribute: Sex}
\label{tbl:result-german-credit}

\begin{tabular}{lccccccccc}\hline 
\textbf{Model} & \textbf{BA} & \textbf{Acc.} & \textbf{SP} & \textbf{EO} & \textbf{EOd} & \textbf{PP} & \textbf{PE} & \textbf{TE} & \textbf{ABROCA} \\ \hline
\textbf{DT} & 0.5954 & 0.6567 & 0.0485 & 0.0160 & 0.1807 & 0.0292 & 0.1646 & 0.0769 & 0.0903 \\
\textbf{NB} & \textbf{0.6604} & \textbf{0.7300} & \underline{0.0019} & 0.0614 & 0.1615 & 0.0166 & 0.1001 & -0.2557 & 0.1012 \\
\textbf{MLP} & 0.6095 & 0.6634 & -0.0669 & 0.0936 & 0.1292 & 0.0214 & 0.0356 & -0.6250 & 0.0697 \\
\textbf{kNN} & 0.5348 & 0.6500 & 0.0641 & 0.0670 & 0.1171 & 0.0391 & 0.0501 & 0.1399 & 0.0458 \\\hdashline
\textbf{DIR-DT} & 0.6221 & 0.6767 & -0.0736 & 0.0972 & 0.1489 & 0.0263 & 0.0517 & -0.6653 & \underline{0.0227} \\
\textbf{DIR-NB} & \underline{0.6392} & 0.7133 & -0.0094 & 0.0511 & 0.0983 & \underline{0.0043} & 0.0473 & -0.2667 & 0.0970 \\
\textbf{DIR-MLP} & 0.5676 & 0.7000 & -0.0326 & 0.0625 & \underline{0.0781} & 0.0169 & \underline{0.0156} & -0.2178 & 0.1114 \\
\textbf{DIR-kNN} & 0.5118 & 0.6267 & -0.0144 & 0.0608 & 0.1431 & \textbf{0.0018} & 0.0823 & -0.2114 & 0.1223 \\
\textbf{LFR-DT} & 0.5686 & 0.5933 & -0.0174 & 0.0646 & 0.1325 & 0.0128 & 0.0679 & -0.3510 & \textbf{0.0032} \\
\textbf{LFR-NB} & 0.4861 & 0.5433 & -0.0592 & 0.0410 & 0.1361 & 0.0546 & 0.0950 & -0.5092 & 0.0342 \\
\textbf{LFR-MLP} & 0.5 & 0.6967 & \textbf{0.0} & \textbf{0.0} & \textbf{0.0} & 0.0371 & \textbf{0.0} & \textbf{0.0} & 0.0545 \\
\textbf{LFR-kNN} & 0.5467 & 0.6667 & -0.0570 & 0.0649 & 0.1161 & 0.0336 & 0.0512 & -0.2826 & 0.0434 \\\hdashline
\textbf{AdaFair} & 0.5554 & 0.7133 & -0.0442 & 0.0364 & 0.1104 & 0.0453 & 0.0740 & -0.1184  & 0.0801 \\
\textbf{Agarwal's} & 0.6289 & 0.7033 & -0.0945 & 0.1116 & 0.2001 & 0.0350 & 0.0884 & -0.6364 & 0.0384 \\\hdashline
\textbf{EOP-DT} & 0.5954 & 0.6567 & 0.0485 & 0.0160 & 0.1807 & 0.0292 & 0.1646 & 0.0769 & NaN \\
\textbf{EOP-NB} & 0.6286 & 0.6900 & -0.0935 & 0.1347 & 0.1703 & 0.0131 & 0.0356 & -0.7500 & NaN \\
\textbf{EOP-MLP} & 0.5990 & 0.6533 & -0.0877 & 0.1069 & 0.1770 & 0.0297 & 0.0701 & -0.7279 & NaN \\
\textbf{EOP-kNN} & 0.5309 & 0.6533 & 0.0877 & 0.0870 & 0.1693 & 0.0357 & 0.0823 & 0.2190 & NaN \\
\textbf{CEP-DT} & 0.5954 & 0.6567 & 0.0485 & 0.0160 & 0.1807 & 0.0292 & 0.1646 & 0.0769 & NaN \\
\textbf{CEP-NB} & 0.6153 & \underline{0.7233} & 0.1151 & \underline{0.0120} & 0.3218 & 0.0597 & 0.3098 & 0.1980 & NaN \\
\textbf{CEP-MLP} & 0.5667 & 0.6600 & 0.0510 & 0.0136 & 0.1877 & 0.0227 & 0.1741 & \underline{0.0208} & NaN \\
\textbf{CEP-kNN} & 0.5257 & 0.6633 & 0.1396 & 0.1337 & 0.2805 & 0.0307 & 0.1468 & 0.3783 & NaN \\
\hline
\end{tabular}
\end{minipage}}
\end{table}

\vspace{-5pt}

\subsubsection{PAKDD Credit Dataset}
\label{subsubsec:result-pakdd-credit}
\begin{figure*}[h]
\centering
\begin{subfigure}{.30\linewidth}
    \centering
    \includegraphics[width=\linewidth]{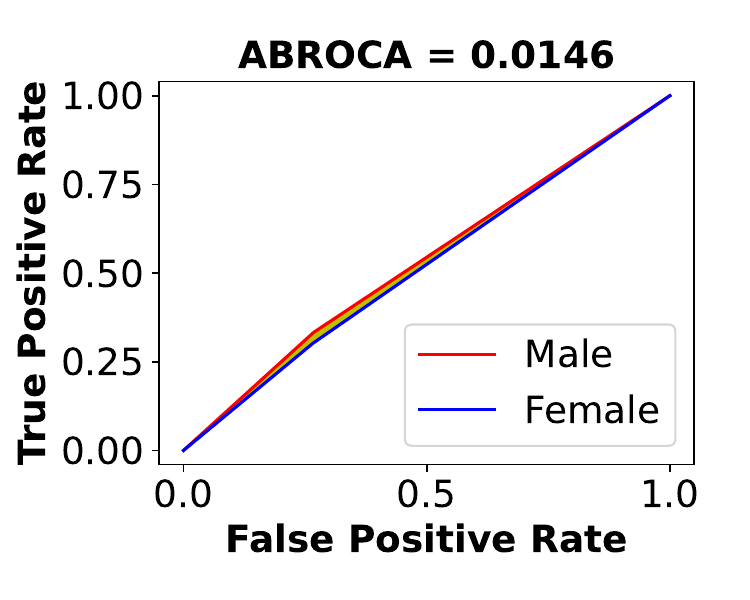}
    \caption{DT}
\end{subfigure}
\hfill
\vspace{-5pt}
\begin{subfigure}{.30\linewidth}
    \centering
    \includegraphics[width=\linewidth]{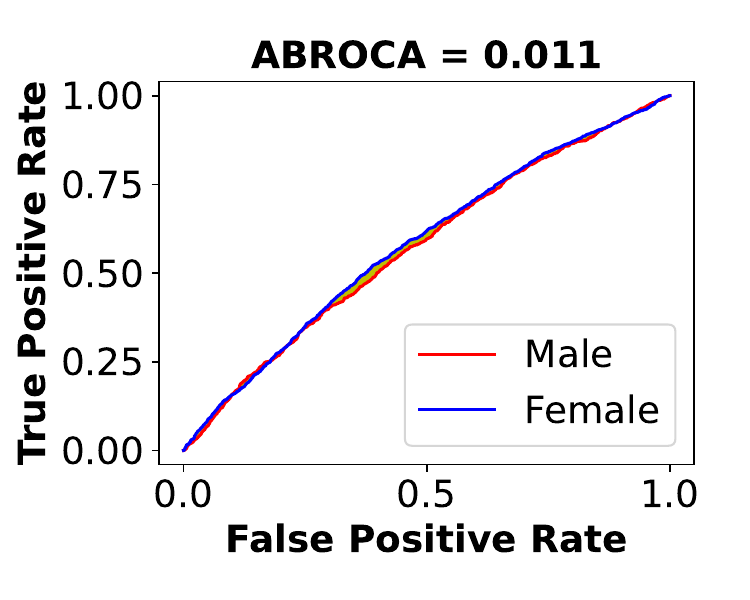}
    \caption{NB}
\end{subfigure}    
\hfill
\vspace{-5pt}
\begin{subfigure}{.30\linewidth}
    \centering
    \includegraphics[width=\linewidth]{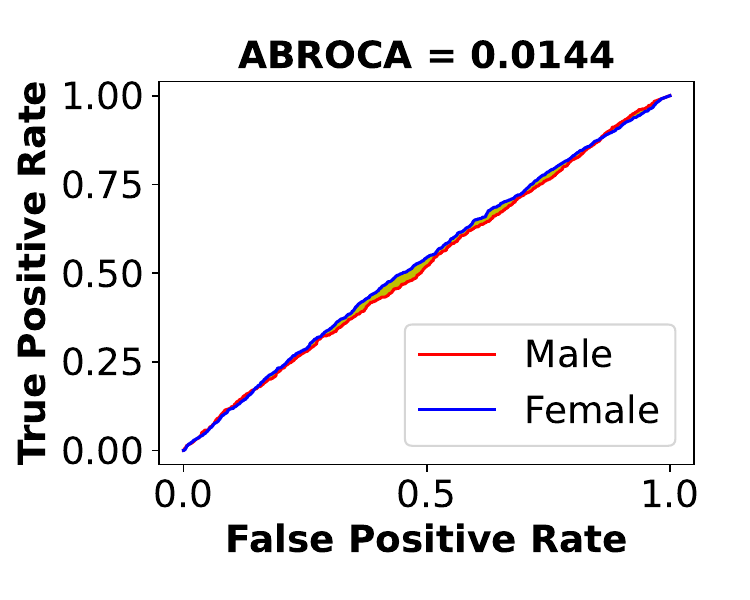}
    \caption{MLP}
\end{subfigure}
\bigskip
\vspace{-5pt}
\begin{subfigure}{.30\linewidth}
    \centering
    \includegraphics[width=\linewidth]{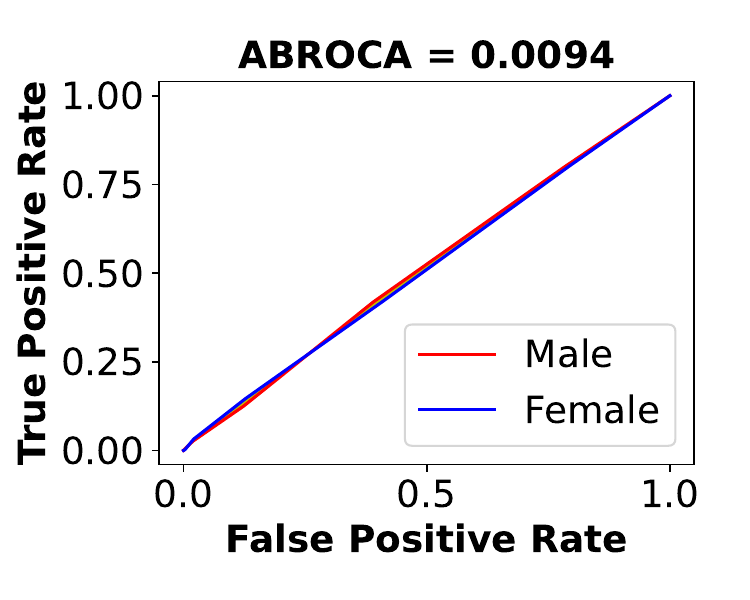}
    \caption{kNN}
\end{subfigure}
\hfill
\vspace{-5pt}
\begin{subfigure}{.30\linewidth}
    \centering
    \includegraphics[width=\linewidth]{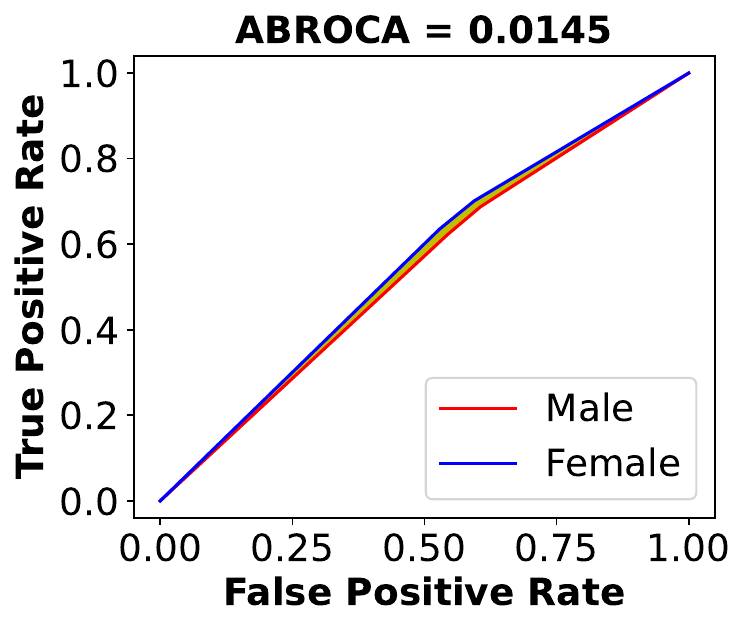}
    \caption{AdaFair}
\end{subfigure}
\vspace{-5pt}
 \hfill
\begin{subfigure}{.30\linewidth}
    \centering
    \includegraphics[width=\linewidth]{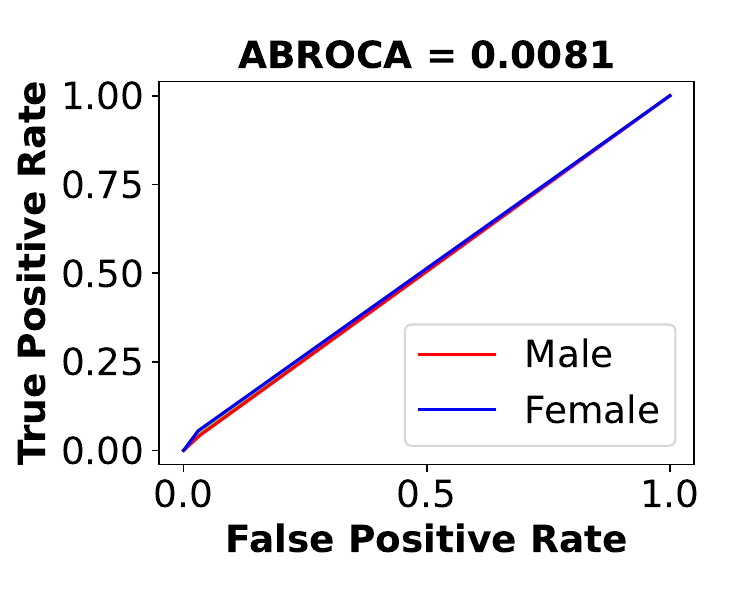}
    \caption{Agarwal's}
\end{subfigure}
\bigskip
\vspace{-5pt}
\begin{subfigure}{.30\linewidth}
    \centering
    \includegraphics[width=\linewidth]{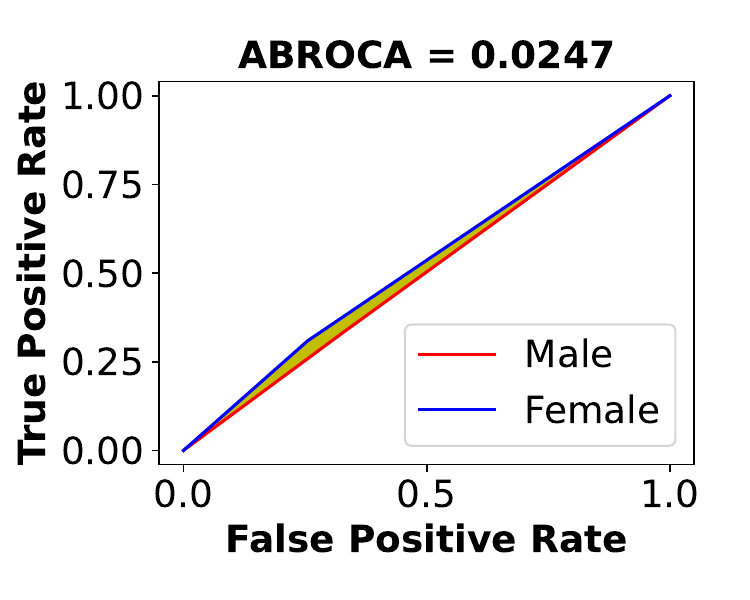}
    \caption{DIR-DT}
\end{subfigure}
\hfill
\vspace{-5pt}
\begin{subfigure}{.30\linewidth}
    \centering
    \includegraphics[width=\linewidth]{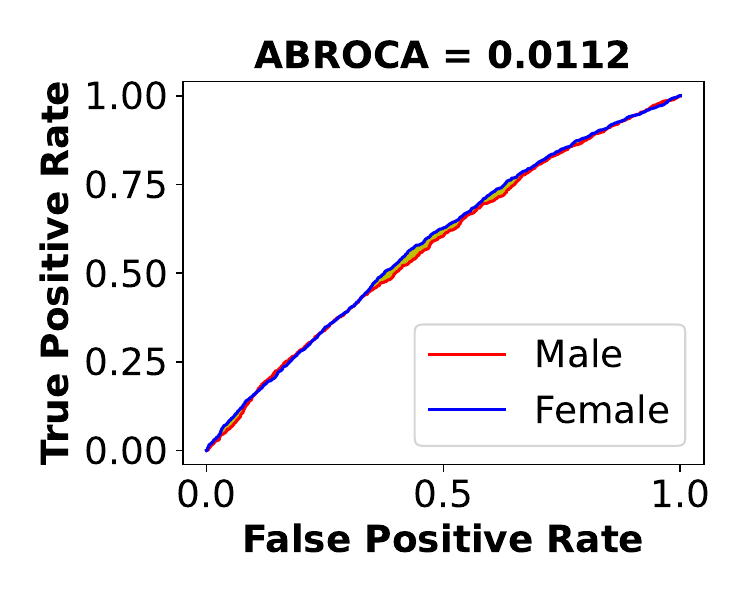}
    \caption{DIR-NB}
\end{subfigure}
\vspace{-5pt}
 \hfill
\begin{subfigure}{.30\linewidth}
    \centering
    \includegraphics[width=\linewidth]{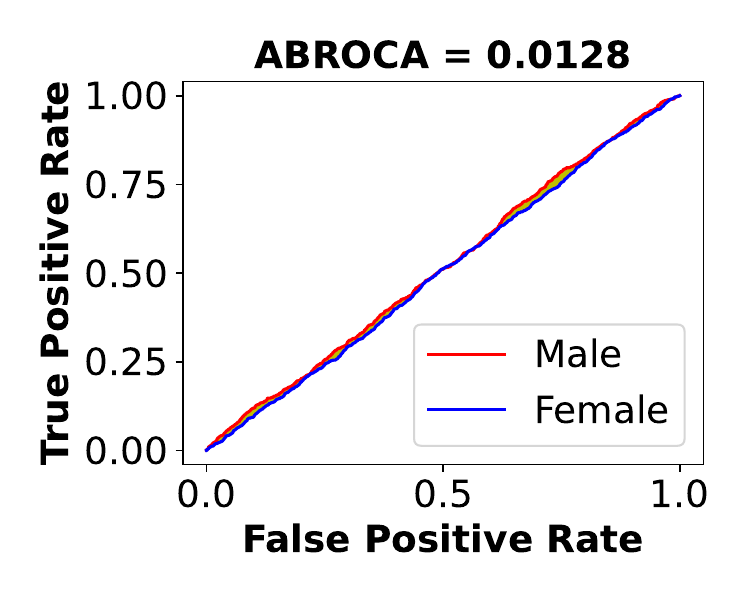}
    \caption{DIR-MLP}
\end{subfigure}
\bigskip
\vspace{-5pt}
\begin{subfigure}{.30\linewidth}
    \centering
    \includegraphics[width=\linewidth]{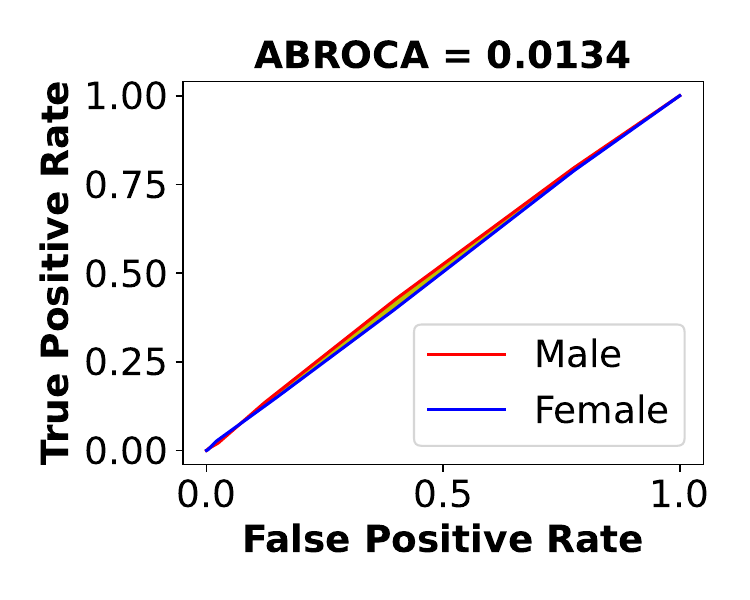}
    \caption{DIR-kNN}
\end{subfigure}
\hfill
\vspace{-5pt}
\begin{subfigure}{.30\linewidth}
    \centering
    \includegraphics[width=\linewidth]{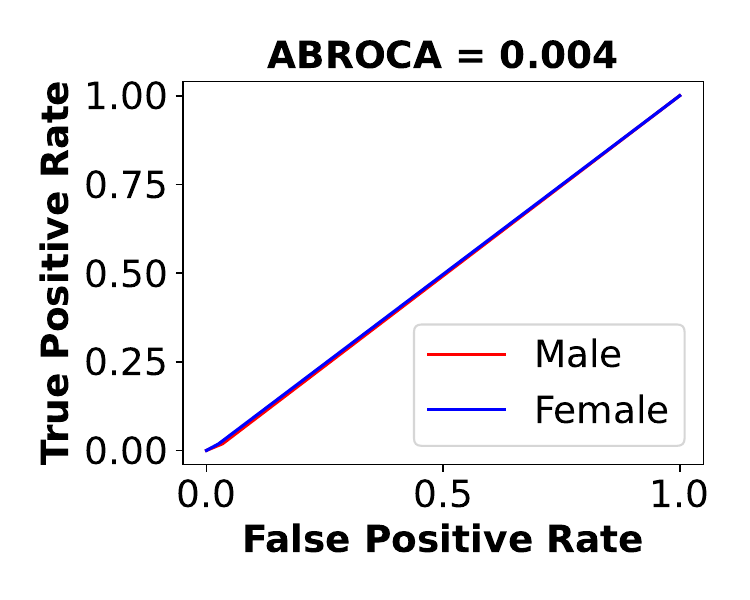}
    \caption{LFR-DT}
\end{subfigure}
\vspace{-5pt}
 \hfill
\begin{subfigure}{.30\linewidth}
    \centering
    \includegraphics[width=\linewidth]{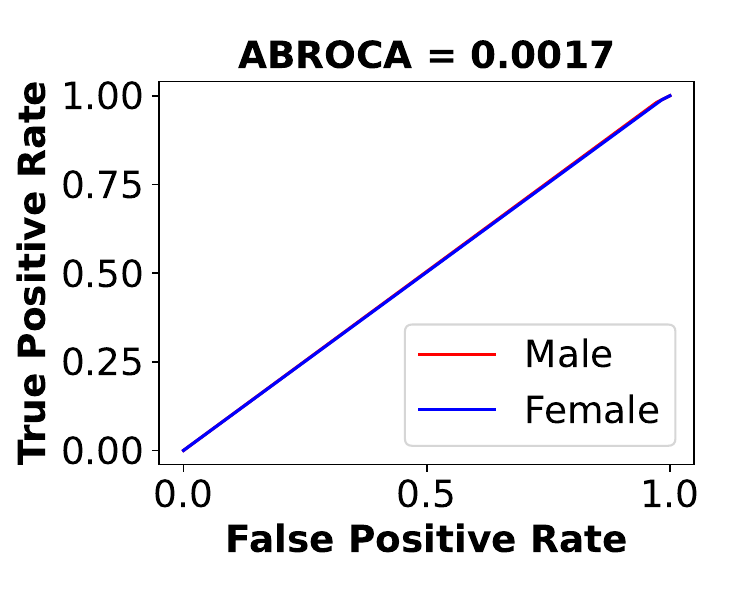}
    \caption{LFR-NB}
\end{subfigure}
\bigskip
\vspace{-5pt}
\begin{subfigure}{.30\linewidth}
    \centering
    \includegraphics[width=\linewidth]{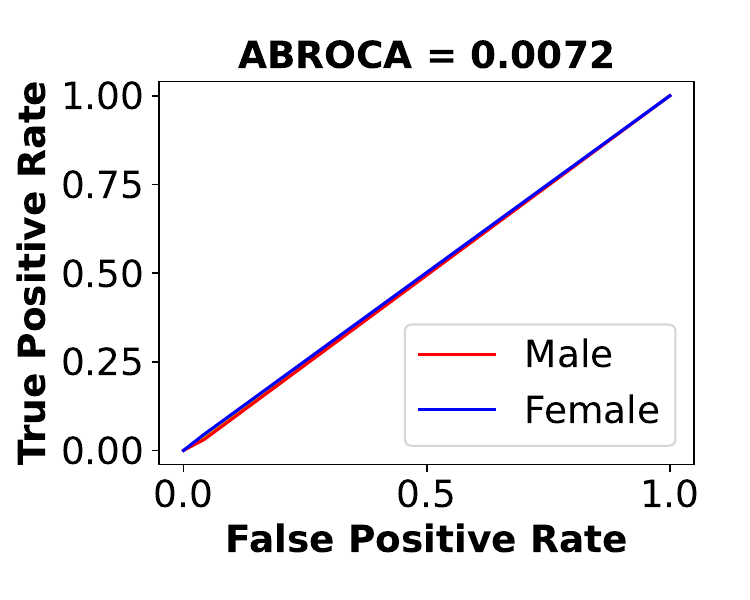}
    \caption{LFR-MLP}
\end{subfigure}
\vspace{-5pt}
\begin{subfigure}{.30\linewidth}
    \centering
    \includegraphics[width=\linewidth]{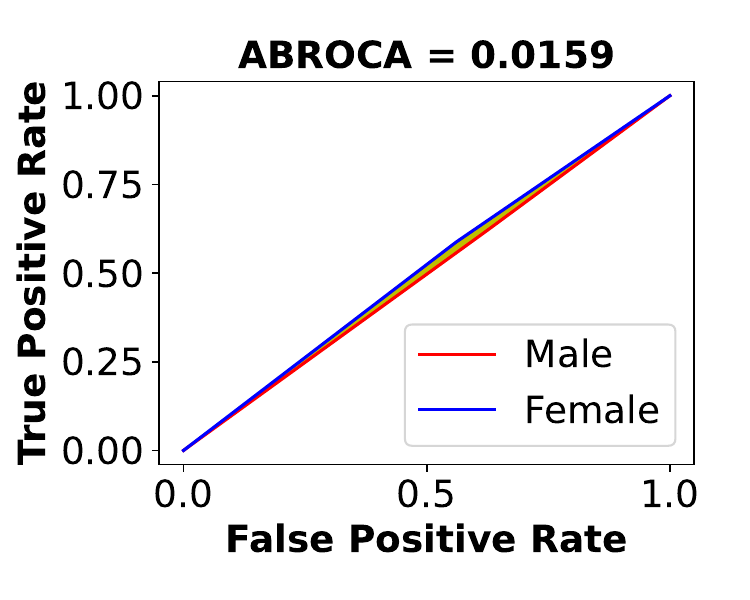}
    \caption{LFR-kNN}
\end{subfigure}
\caption{PAKDD credit: ABROCA slice plots}
\label{fig:pakdd-abroca}
\end{figure*}

In the PAKDD credit risk assessment dataset, all methods exhibit low balanced accuracy, as shown in Table \ref{tbl:result-pakdd} and Figure \ref{fig:pakdd-abroca}. Among the in-processing approaches, AdaFair achieves superior performance, obtaining the best results on 6 out of 9 evaluation measures, including accuracy and five fairness measures. In the pre-processing group, although LFR-NB achieves improved fairness in classification outcomes (as measured by TE and ABROCA), this comes at the cost of a significant reduction in accuracy. In addition, the post-processing models outperform the other approaches in terms of balanced accuracy.

\begin{table}[h]
\resizebox{\textwidth}{!}
{
\begin{minipage}{\textwidth}
\centering
\caption{PAKDD credit: performance of predictive models. Protected attribute: Sex}
\label{tbl:result-pakdd}

\begin{tabular}{lccccccccc}\hline
\textbf{Model} & \textbf{BA} & \textbf{Acc.} & \textbf{SP} & \textbf{EO} & \textbf{EOd} & \textbf{PP} & \textbf{PE} & \textbf{TE} & \textbf{ABROCA} \\ \hline
\textbf{DT} & \textbf{0.5241} & 0.6244 & 0.0124 & 0.0325 & 0.0358 & 0.0476 & 0.0033 & -0.0707 & 0.0146 \\
\textbf{NB} & 0.5088 & 0.7256 & 0.0022 & 0.0087 & 0.0143 & 0.0523 & 0.0056 & 0.3997 & 0.0110 \\
\textbf{MLP} & 0.5119 & 0.6925 & 0.0655 & 0.0715 & 0.1340 & 0.0224 & 0.0624 & 1.5546 & 0.0144 \\
\textbf{kNN} & 0.5057 & 0.6822 & -0.0056 & 0.0192 & 0.0201 & \underline{0.0013} & 0.0010 & -0.4404 & 0.0094 \\\hdashline
\textbf{DIR-DT} & \underline{0.5174} & 0.6189 & 0.0253 & 0.0116 & 0.0493 & 0.0090 & 0.0377 & \underline{-0.0109} & 0.0247 \\
\textbf{DIR-NB} & 0.5130 & 0.7210 & 0.0251 & 0.0155 & 0.0432 & 0.0556 & 0.0277 & 3.4949 & 0.0112 \\
\textbf{DIR-MLP} & 0.5003 & \underline{0.7351} & \textbf{0.0} & 0.0018 & \underline{0.0024} & 0.5833 & \underline{0.0006} & -909.4 & 0.0128 \\
\textbf{DIR-kNN} & 0.5027 & 0.6810 & -0.0028 & 0.0044 & 0.0101 & 0.0437 & 0.0056 & -0.4816 & 0.0134 \\
\textbf{LFR-DT} & 0.4949 & 0.7200 & 0.0069 & \underline{0.0013} & 0.0106 & 0.0189 & 0.0093 & 2.6914 & \underline{0.0040} \\
\textbf{LFR-NB} & 0.5034 & 0.2771 & -0.0088 & 0.0064 & 0.0163 & 0.0271 & 0.0100 & \textbf{-0.0033} & \textbf{0.0017} \\
\textbf{LFR-MLP} & 0.4986 & 0.7180 & 0.0034 & 0.0070 & 0.0145 & 0.0543 & 0.0074 & 0.6436 & 0.0072 \\
\textbf{LFR-kNN} & 0.5075 & 0.4736 & 0.0212 & 0.0024 & 0.0319 & 0.0155 & 0.0294 & -0.0239 & 0.0159 \\\hdashline
\textbf{AdaFair} & 0.5 & \textbf{0.7353} & \textbf{0.0} & \textbf{0.0} & \textbf{0.0} & \textbf{0.0} & \textbf{0.0} & NaN & 0.0145 \\
\textbf{Agarwal's} & 0.5093 & 0.7263 & -0.0017 & 0.0140 & 0.0162 & 0.0549 & 0.0021 & -0.9031 & 0.0081 \\\hdashline
\textbf{EOP-DT} & \textbf{0.5241} & 0.6244 & 0.0124 & 0.0325 & 0.0358 & 0.0476 & 0.0033 & -0.0707 & NaN \\
\textbf{EOP-NB} & 0.5083 & 0.7258 & \underline{-0.0009} & 0.0134 & 0.0165 & 0.0613 & 0.0031 & -0.5122 & NaN \\
\textbf{EOP-MLP} & 0.5132 & 0.6854 & 0.0120 & 0.0123 & 0.0233 & 0.0270 & 0.0110 & -0.0755 & NaN \\
\textbf{EOP-kNN} & 0.5057 & 0.6817 & -0.0076 & 0.0214 & 0.0242 & \underline{0.0013} & 0.0029 & -0.4848 & NaN \\
\textbf{CEP-DT} & \textbf{0.5241} & 0.6244 & 0.0124 & 0.0325 & 0.0358 & 0.0476 & 0.0033 & -0.0707 & NaN \\
\textbf{CEP-NB} & 0.5088 & 0.7269 & 0.0068 & 0.0037 & 0.0138 & 0.0696 & 0.0101 & 2.6232 & NaN \\
\textbf{CEP-MLP} & 0.5127 & 0.7029 & 0.0996 & 0.1051 & 0.2018 & 0.0069 & 0.0967 & 4.6288 & NaN \\
\textbf{CEP-kNN} & 0.5060 & 0.6860 & 0.0069 & 0.0070 & 0.0186 & 0.0053 & 0.0116 & -0.1113 & NaN \\
\hline
\end{tabular}
\end{minipage}}
\end{table}

\vspace{-5pt}

\subsection{Discussion and Limitation}
\label{subsec:discussion}

In summary, fairness-aware models have achieved good results by balancing model accuracy and fairness in the outcomes. As expected, fairness-aware models achieve the best values with respect to fairness measures. Among these, LFR-MLP and AdaFair are notable methods that outperform others across multiple datasets. Interestingly, AdaFair demonstrates outstanding capability by improving not only accuracy but also fairness in the results. This performance is evidenced in all 5 datasets. Furthermore, the difference in accuracy and balanced accuracy between traditional classification models and fairness-aware methods is not significant.

The paper has some limitations, which provide opportunities for further research. First, the evaluation of fairness in this study is limited to individual protected attributes and does not yet consider the simultaneous impact of multiple protected attributes, such as gender and race, nor the relationships among different fairness measures as well as the intersection fairness. Second, our analysis relies on commonly used real-world datasets, which may not fully capture complex or domain-specific biases, particularly in financial applications where high-quality and unbiased datasets are difficult to obtain. Finally, while this study focuses on fairness assessment, it does not explicitly address the development of fair or explainable classification models, limiting insights into the underlying sources of bias within both the learning algorithms and the datasets.

\section{Conclusions and Outlook}
\label{sec:conclusion}
In this work, we investigated the prevalent credit scoring datasets used in ML for the finance domain. Data analysis using a Bayesian network reveals that bias naturally exists in all the selected datasets, indicating a potential bias in the outcomes of predictive models. Furthermore, we evaluated traditional classifiers in comparison with various fairness-aware models across three approaches: pre-processing, in-processing, and post-processing. The experimental results show that the application of fairness-aware methods demonstrates an improvement in meeting both fairness and accuracy criteria compared to traditional models.

In the future, we plan to expand the evaluation of fairness to simultaneously address multiple protected attributes, such as gender and race, while further exploring the correlations between different fairness measures. Additionally, understanding commonly used datasets motivates us to research and develop fair synthetic data generation models for finance domain, as finding a perfect dataset in the real world has never been a straightforward task. Furthermore, we aim to develop fair and explainable classification algorithms to understand the root causes of bias within the algorithms themselves as well as in the datasets.


\section*{Declarations}
\begin{itemize}
\item \textbf{Competing Interests}
The authors declare that they have no competing interests.

\item \textbf{Funding}
The authors did not receive support from any organization for the submitted work.

\item \textbf{Code availability} 
The source code and dataset are publicly available at \url{https://github.com/tailequy/faircredit}

\item \textbf{Author contribution}
All authors contributed to the study conception and design. Material preparation, data collection and analysis were performed by Huyen Giang Thi Thu, Thang Viet Doan and Tai Le Quy. The first draft of the manuscript was written by Huyen Giang Thi Thu, Ha-Bang Ban and Tai Le Quy and all authors commented on previous versions of the manuscript. All authors read and approved the final manuscript.
\end{itemize}

\bibliography{sn-bibliography}

\end{document}